\documentclass[letterpaper]{article} 
\usepackage{aaai2027}    
\nocopyright
\usepackage[hyphens]{url}            
\usepackage{graphicx}                
\usepackage{natbib}                  
\usepackage{caption}                 
\usepackage{booktabs}
\usepackage{multirow}
\usepackage{mathtools}
\usepackage{amssymb}
\usepackage{pifont}

\usepackage{color}
\usepackage{siunitx} 

\iftrue 

\newcommand{\todop}[1]{\textcolor{red}{\emph{\bf \{#1\} }}}
\newcommand{\comment}[1]{\todop{See comment}}
\else

\newcommand{\todop}[1]{\textcolor{red}{\emph{\bf \{#1\} }}}
\newcommand{\comment}[1]{\todop{See comment}}
\fi

\newcommand{\cmark}{\ding{51}}
\newcommand{\xmark}{\ding{55}}

\begin{document}

\title{
\fontsize{16}{18}\selectfont
From Routes to Steps: Separating Semantic Progress\\[-1mm]
from Local Execution in Vision-and-Language Navigation
}

\author{
\fontsize{10.5}{12}\selectfont
Xiangyun Huang\textsuperscript{1},
Xiangchen Wang\textsuperscript{2},
Runfeng Lin\textsuperscript{1,3},
Yihao Xu\textsuperscript{1},\\
Kangyu Huang\textsuperscript{4},
Jiang Hengchen\textsuperscript{1,5},
Xiwang Dong\textsuperscript{1},
Lin Jiarong\textsuperscript{1,*}
}

\affiliations{
\footnotesize
\textsuperscript{1}Beihang University
\quad
\textsuperscript{2}Southern University of Science and Technology
\quad
\textsuperscript{3}Central South University
\\
\textsuperscript{4}Harbin Institute of Technology, Shenzhen
\quad
\textsuperscript{5}Dalian University of Technology
\\

\small\texttt{23231088@buaa.edu.cn, zivlin@buaa.edu.cn}
}

\maketitle

\begingroup
\renewcommand\thefootnote{}
\footnotetext[1]{* Corresponding author}
\endgroup

\begin{abstract}

Vision-and-Language Navigation (VLN) requires an agent to follow a route-level instruction by executing its constituent steps from egocentric visual observations. Existing VLM-based navigators typically supervise both capabilities through next-action prediction alone, making progress-tracking errors difficult to distinguish from execution errors. When an agent deviates from the route, a corrective action label may recover the next movement but does not indicate whether the agent selected the wrong sub-instruction or failed to execute the correct one. Consequently, the agent may continue making decisions from an erroneous progress state.
To resolve this ambiguity, we propose \textbf{Route2Step}, a framework that decouples semantic progress tracking from action generation through an explicit step-level interface.
The Instruction Analysis Module ($\mathcal{M}_{\mathrm{IA}}$) predicts this state from the global instruction and visual history. Conditioned on the predicted state and recent observations, the Action Generation Module ($\mathcal{M}_{\mathrm{AG}}$) generates local action chunks. 
To supervise the progress state without manual temporal labels, E-SPA, a step-alignment procedure, associates sub-instructions with their corresponding portions of route-level demonstrations. These alignments enable state supervision for incorrect progress estimates, while direct action supervision is reserved for rollout groups that repeatedly fail under the correct active sub-instruction. On R2R-CE, Route2Step improves SR from 48.1\% to 55.3\% and SPL from 43.3\% to 48.2\%,
using 190K state-level corrective samples while requiring only 11.5K directly action-supervised states.
Experiments in real-world indoor and outdoor environments further demonstrate the practical applicability of Route2Step.
The project page is:
\url{https://sisyphus-hxy.github.io/Route2Step/}.
\end{abstract}

\section{Introduction}

Vision-and-Language Navigation (VLN) requires an embodied agent to follow a natural-language route instruction in a visually grounded environment. In continuous VLN-CE benchmarks~\citep{krantz2020beyond,ku2020room}, this requires both identifying the route step currently in progress and executing that step with low-level actions. These capabilities rely on different evidence: semantic progress must be grounded in the global instruction and accumulated visual history, whereas local execution depends primarily on recent observations. 
Under on-path conditions, these two processes are usually consistent and can be learned jointly. However, this consistency breaks once the agent deviates from the demonstrated route, making it necessary to distinguish an incorrect estimate of route progress from a failure to execute the correct step.

\begin{figure}[t]
    \centering
    \includegraphics[width=\columnwidth]{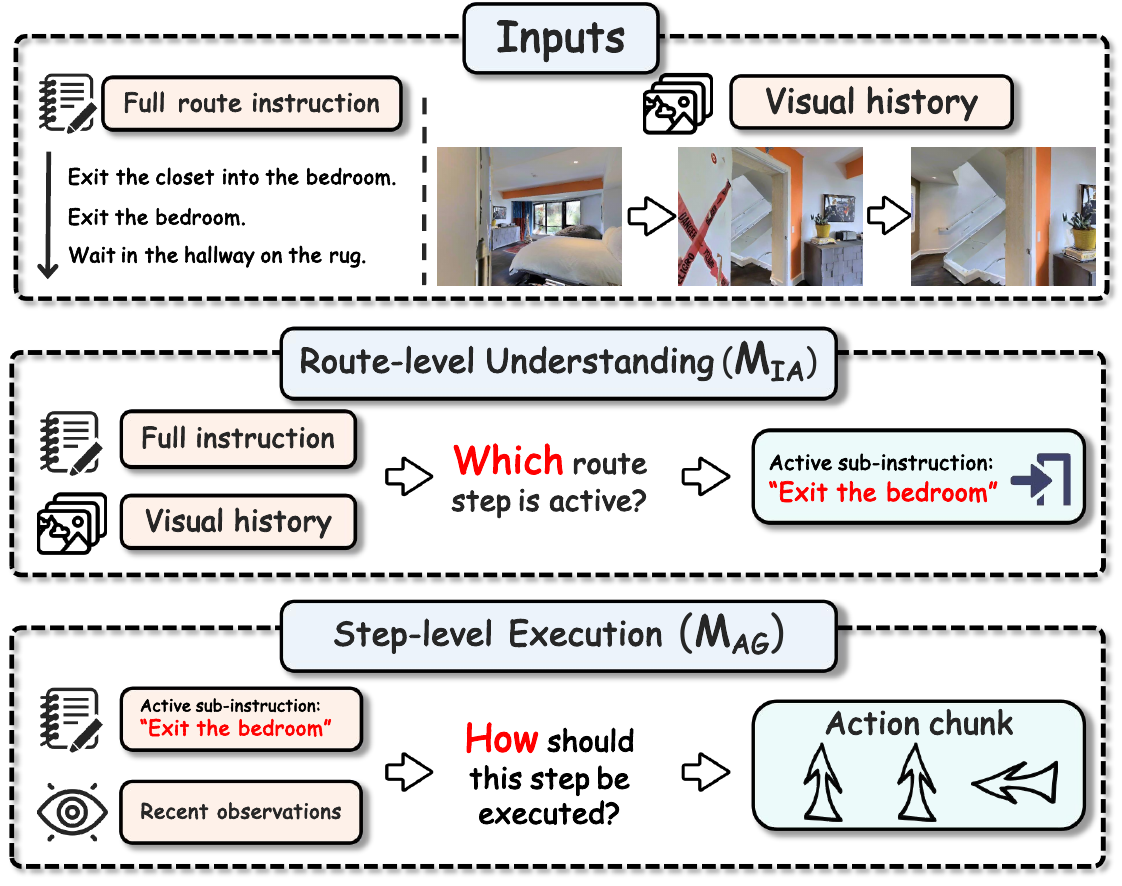}
    \caption{
    Route2Step separates route-level understanding from step-level
    execution. $\mathcal{M}_{\mathrm{IA}}$ identifies the active sub-instruction from the full
    route instruction and visual history, while $\mathcal{M}_{\mathrm{AG}}$ executes it using
    recent observations.
    }
    \label{fig:teaser}
    \vspace{-4mm}
\end{figure}

Recent VLM-based navigators have improved continuous VLN through pretrained multimodal representations and large-scale action supervision~\citep{navid,streamvln}. Most remain unified policies that map language and visual context directly to low-level actions. Although some introduce intermediate reasoning or progress signals~\citep{wang2025auxthink,awarevln,wang2025progressthink}, these signals are usually learned jointly with action generation. Consequently, the training objective conflates a \emph{semantic-progress error}, in which the wrong sub-instruction is active, with a \emph{local-execution error}, in which the correct sub-instruction is executed poorly. This conflation is amplified by DAgger-style post-training~\citep{ross2011dagger}, which relabels policy-induced states with expert next actions. At an off-path state, such labels can correct a local motion but do not separately supervise whether the active sub-instruction itself should change. Action training may therefore improve short-term recovery while leaving semantic progress implicit and misaligned for subsequent decisions.

To address this coupling, we propose Route2Step, a level-separated framework that makes route progress an explicit interface between semantic tracking and local action generation. As illustrated in Fig.~\ref{fig:teaser}, the Instruction Analysis Module ($\mathcal{M}_{\mathrm{IA}}$) receives the global instruction and visual history and predicts the active sub-instruction. It additionally predicts an execution state (\textsc{Normal} or \textsc{Recovering}). The Action Generation Module ($\mathcal{M}_{\mathrm{AG}}$) then uses this interface and a recent observation window to generate local action chunks. In other words, $\mathcal{M}_{\mathrm{IA}}$ determines \emph{which} route step is active and whether it should be pursued normally or through recovery, while $\mathcal{M}_{\mathrm{AG}}$ determines \emph{how} to execute that step from recent observations. The two modules are trained independently and operate over their respective temporal contexts, preventing action-level updates from directly overwriting the semantic-progress estimator.

This explicit interface also determines how Route2Step uses policy-induced experience. We first align each instruction with step-level trajectory segments and associate every sub-instruction with a semantic waypoint. During rollouts, the active sub-instruction is retained until its waypoint is reached, even when the agent deviates from the expert path. These rollout histories therefore provide state-level supervision for $\mathcal{M}_{\mathrm{IA}}$ to maintain or update semantic progress and to predict the execution state. Direct expert-action labels are introduced for $\mathcal{M}_{\mathrm{AG}}$ only when repeated attempts fail to complete the same active step; they improve local recovery without changing the semantic target. Route2Step consequently assigns supervision to the level it is intended to improve, rather than treating every off-path state as an action-prediction error.

Step-level learning requires sub-instruction-trajectory supervision. For RxR, we map the landmark-level annotations from discrete Landmark-RxR~\citep{he2021landmark} to the corresponding continuous RxR-CE trajectories. For R2R, where standard demonstrations contain only route-level instructions, we introduce \textbf{E-SPA} (Energy-minimizing Semantic Path Alignment), an offline procedure that aligns fine-grained, locally executable sub-instructions with contiguous trajectory segments. Human-annotated R2R sub-instruction--trajectory alignments from Fine-Grained R2R (FG-R2R) are excluded from training and reserved solely for external evaluation~\citep{hong2020FG_R2R}. The resulting alignments support semantic-progress prediction in $\mathcal{M}_{\mathrm{IA}}$ and step-conditioned action learning in $\mathcal{M}_{\mathrm{AG}}$.

Experiments validate both the navigation benefit and the supervision allocation. On R2R-CE, Route2Step improves success rate (SR) from 48.1\% to 55.3\% and success weighted by path length (SPL) from 43.3\% to 48.2\%. It uses approximately 190K rollout samples with state-level supervision, but only 11.5K directly action-supervised rollout states. Controlled off-path evaluation further shows substantially stronger active-sub-instruction tracking, while results on RxR-CE, plug-and-play transfer to existing VLN policies, and physical-robot deployments demonstrate the robustness and practical value of the semantic--execution interface.

Our contributions are threefold:
\begin{itemize}
    \item We reformulate policy-induced corrective learning in VLN as a level-separated problem: state-level supervision corrects semantic progress, while selective action-level supervision addresses local execution recovery.

    \item We propose Route2Step, a dual-module framework with an explicit semantic--execution interface, and introduce E-SPA to derive step-level sub-instruction--trajectory alignments from route-level R2R demonstrations without additional manual temporal annotation.

    \item We show that level-separated correction improves R2R-CE SR from 48.1\% to 55.3\% while using only 11.5K directly action-supervised rollout states. Evaluations on controlled off-path histories, RxR-CE, transfer to existing VLN policies, and physical robots further support the effectiveness and transferability of the proposed interface.
\end{itemize} 

\begin{figure*}[tbp]
    \centering
    \includegraphics[width=\textwidth]{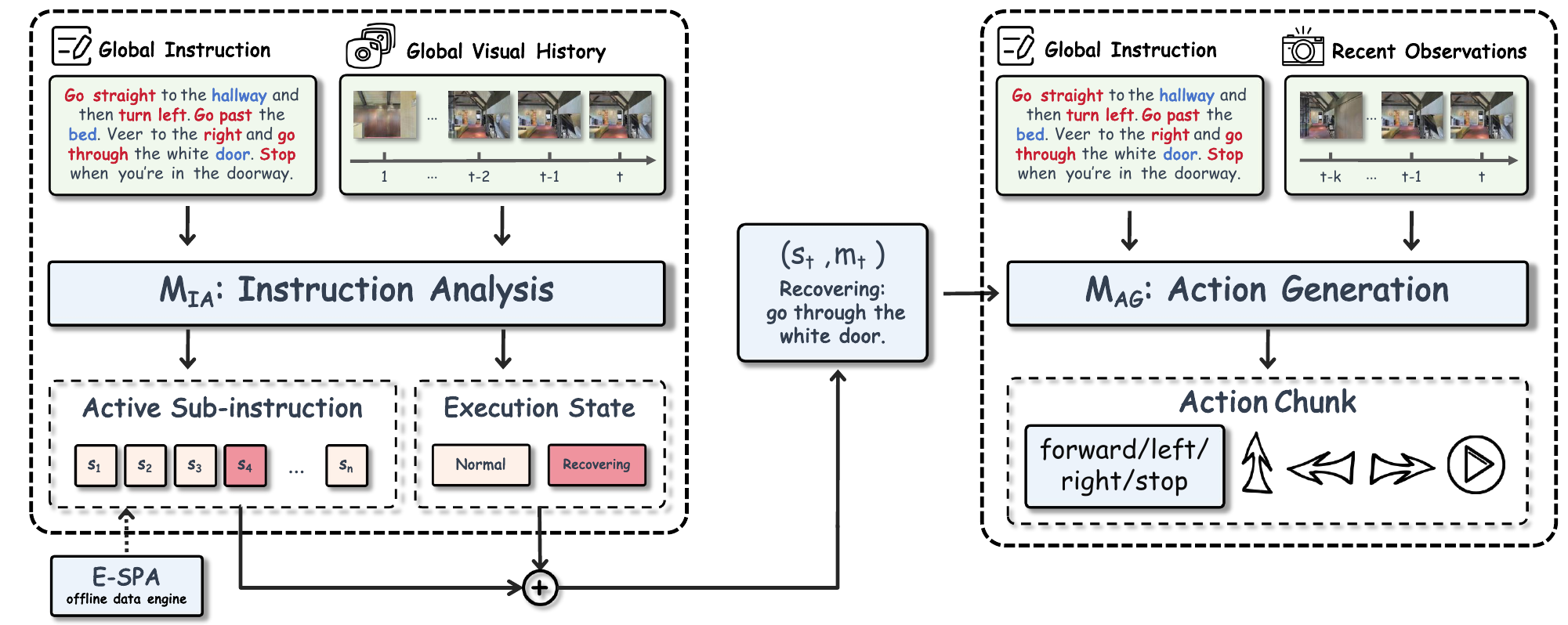}
    \caption{
    Overview of Route2Step.
    Given the global instruction and visual history,
    $\mathcal{M}_{\mathrm{IA}}$ predicts the active sub-instruction and
    execution state, identifying which route step remains active and whether
    execution is \textsc{Normal} or \textsc{Recovering}.
    Conditioned on the global instruction, this explicit interface, and
    recent observations, $\mathcal{M}_{\mathrm{AG}}$ predicts 
    action chunks for executing the active step.
    }
    \label{fig:framework}
    \vspace{-2mm}
\end{figure*}

\section{Related Work}

\subsection{VLM-based VLN and Corrective Learning}

Recent VLN-CE methods increasingly adapt pretrained VLMs to predict low-level actions directly from language instructions and egocentric visual histories~\citep{navid,uninavid,streamvln,zhu2026textscnavidavisionlanguagenavigationinverse,wang2026livevlnbreakingstopandgoloop}.
Although these models benefit from large-scale imitation learning, they remain vulnerable to policy-induced states absent from expert demonstrations.
Recent approaches address this issue through reinforcement post-training or corrective data construction.
For example, ActiveVLN introduces multi-turn reinforcement learning, whereas BudVLN constructs semantically consistent corrective trajectories~\citep{zhang2025activevln,budvln}. 
CorrectNav constructs action- and perception-oriented self-correction data, Efficient-VLN reduces DAgger exploration overhead through a dynamic mixed
policy, and DecoVLN performs state-action corrective finetuning with decoupled observation and reasoning~\citep{correctnav,efficientvln,decovln}.
Although these methods differ in mechanism, they primarily use policy-induced experience to improve the action policy.
Route2Step instead uses policy-induced histories for semantic-progress correction while selectively introducing expert actions for local recovery.

\subsection{Progress Modeling and Step-Level Supervision}

Earlier VLN agents used scalar progress estimators to monitor trajectory completion or trigger heuristic backtracking ~\citep{ma2019selfmonitoring,ma2019regretful}.
More recent studies expose intermediate reasoning or semantic-progress signals for long-horizon navigation. Aux-Think uses structured chain-of-thought as auxiliary supervision for a unified action policy~\citep{wang2025auxthink}, while AwareVLN integrates completion, deviation, and stopping awareness into a unified reason--act policy~\citep{awarevln}.
Progress-Think instead employs a dedicated progress module to guide a separate navigation policy, followed by joint progress--policy co-finetuning~\citep{wang2025progressthink}.
Related hierarchical methods connect high-level planning and low-level execution through language actions or spatial goals~\citep{cheng2025navila,wei2025groundslowfastdualsystem}.
Route2Step instead assigns semantic-progress estimation and local execution distinct temporal contexts, optimization objectives, and corrective supervision, while coupling them through an explicit interface comprising the active sub-instruction and execution state.
Fine-grained semantic-progress supervision further requires instruction-trajectory alignment. 
Landmark-RxR provides human-annotated landmark-level alignments, while automatic methods associate instruction semantics with trajectory observations~\citep{he2021landmark,song2024fine}. 
For R2R, E-SPA constructs this supervision from route-level demonstrations by producing locally executable sub-instruction--trajectory pairs without additional manual temporal annotation.

\section{Method}
\label{sec:method}

Route2Step separates semantic-progress estimation from local action
execution through an explicit interface comprising the active
sub-instruction and execution state (Fig.~\ref{fig:framework}).

\subsection{Semantic--Execution Interface}
\label{sec:framework}

Given a global instruction $\mathcal{I}$, the agent observes an egocentric visual stream $V_{1:t}$ and predicts a short chunk of low-level navigation actions $a_{t:t+h}$. Instead of directly modeling $P(a_{t:t+h}\mid\mathcal{I},V_{1:t})$, Route2Step introduces two independently trainable modules:
\begin{align}
(s_t,m_t) &=
\mathcal{M}_{\mathrm{IA}}(\mathcal{I},V_{1:t}), \label{eq:mia}\\
a_{t:t+h} &=
\mathcal{M}_{\mathrm{AG}}
(\mathcal{I},s_t,m_t,V_{t-k:t}), \label{eq:mag}
\end{align}
where $s_t$ is the active sub-instruction and
$m_t\in\{\textsc{Normal},\textsc{Recovering}\}$ denotes the execution
state.

The instruction-analysis module $\mathcal{M}_{\mathrm{IA}}$ uses the global visual history to determine which semantic step currently remains active. The action-generation module $\mathcal{M}_{\mathrm{AG}}$ operates on a short recent observation window and predicts local actions conditioned on the interface. Here, $s_t$ specifies \emph{what} should currently be accomplished, whereas $m_t$ indicates whether it should be executed nominally or from a recovery state. The global instruction is retained by $\mathcal{M}_{\mathrm{AG}}$ to preserve task context and stopping semantics.

\subsection{Step-Level Supervision Construction}
\label{sec:step_supervision}
Training Route2Step requires aligned sub-instruction--trajectory pairs that ground semantic steps in physical trajectories. To construct these aligned pairs for R2R, E-SPA first rewrites each route instruction into an ordered sequence of executable sub-instructions $S=\{s_1,\ldots,s_n\}$ while preserving the original semantic order. It then partitions an expert trajectory of length $T$ into $n$ contiguous segments with boundaries $B=\{b_1,\ldots,b_{n+1}\}$, where $b_1=1$ and $b_{n+1}=T+1$.

For a candidate assignment of the $k$-th sub-instruction $s_k$ to a non-empty frame interval $[i,j]$, where $1 \leq i \leq j \leq T$, we define
\begin{equation}
\begin{aligned}
C(s_k,i,j) = {}&
\lambda_{\mathrm{sem}} C_{\mathrm{sem}}(s_k,i,j)
+
\lambda_{\mathrm{act}} C_{\mathrm{act}}(s_k,i,j) \\
&+
\lambda_{\mathrm{dur}} C_{\mathrm{dur}}(i,j)
+
\lambda_{\mathrm{anchor}} C_{\mathrm{anchor}}(s_k,i,j).
\end{aligned}
\end{equation}
The non-negative coefficients $\lambda_{\mathrm{sem}}$, $\lambda_{\mathrm{act}}$, $\lambda_{\mathrm{dur}}$, and $\lambda_{\mathrm{anchor}}$ balance the corresponding costs. 
Here, $C_{\mathrm{sem}}$ measures visual--text compatibility between the sub-instruction and the trajectory segment~\citep{radford2021clip}, $C_{\mathrm{act}}$ measures consistency between linguistic motion intent and executed actions, and $C_{\mathrm{dur}}$ discourages implausibly short or imbalanced segments. $C_{\mathrm{anchor}}$ incorporates trajectory-aware anchoring cues, particularly height-changing intervals associated with stair-related sub-instructions, to reduce ambiguous alignments.
The implementation of the anchoring cues and the complete cost definitions are provided in the supplementary material.

Following classical monotonic sequence alignment \citep{sakoe1978dynamic} and ordered video--text alignment \citep{bojanowski2015weakly}, the optimal boundaries are obtained by dynamic programming:
\begin{equation}
B^*
=
\arg\min_{1=b_1<\cdots<b_{n+1}=T+1}
\sum_{k=1}^{n}
C(s_k,b_k,b_{k+1}-1).
\end{equation}
This produces temporally aligned (sub-instruction, trajectory-segment) pairs used to supervise both active sub-instruction prediction in $\mathcal{M}_{\mathrm{IA}}$ and local execution in $\mathcal{M}_{\mathrm{AG}}$. For each sub-instruction $s_k$, we map the endpoint frame $b_{k+1}-1$ of its aligned segment to the corresponding ground-truth pose in the Habitat coordinate system~\citep{savva2019habitat}. The resulting semantic waypoint retains both position and heading and serves as a pose-level completion criterion for policy-induced trajectories.

\begin{figure}[t]
    \centering
    \includegraphics[width=\columnwidth]{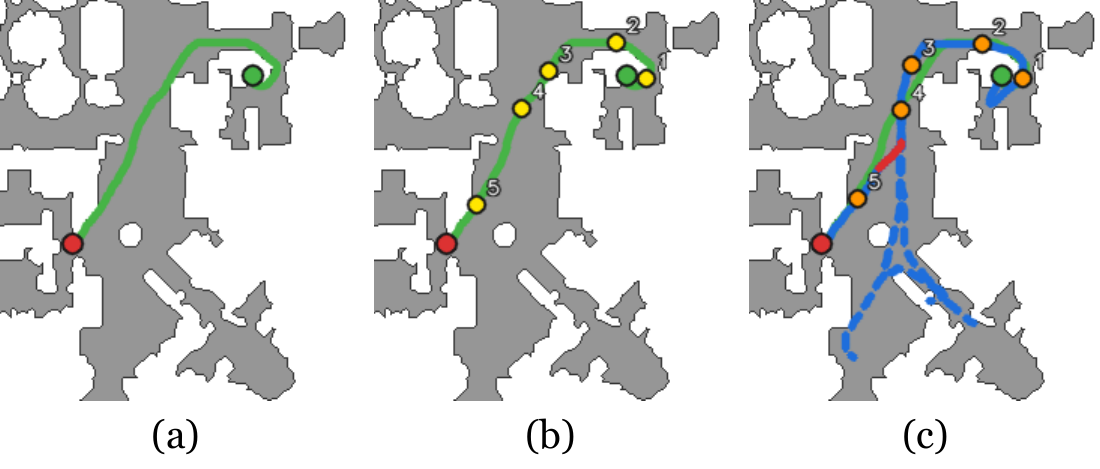}
    \caption{
    Geometry-grounded supervision.
    (a) Expert path.
    (b) Step-aligned semantic waypoints.
    (c) Fixed-sub-instruction rollout with expert-guided recovery.
    \protect\newline
    Dot colors: red/green for start/goal and yellow/orange for waypoints.
   Path styles: solid blue/red for \textsc{Normal}/\textsc{Recovering}, and dashed blue for failed rollouts.
    }
    \label{fig:dagger_topdown}
    \vspace{-3mm}
\end{figure}

For RxR, we map the landmark-level annotations from discrete Landmark-RxR onto the corresponding RxR-CE trajectories and use them as step-level training supervision~\citep{he2021landmark}. For R2R evaluation, we map the human-annotated sub-instruction--path alignments from Fine-Grained R2R (FG-R2R) onto the corresponding R2R-CE trajectories~\citep{hong2020FG_R2R}. These FG-R2R annotations are excluded from training and used only for direct evaluation of $\mathcal{M}_{\mathrm{IA}}$. The mapping procedure and correspondence rules are provided in the supplementary material.

\subsection{Level-Separated Training}
\label{sec:training}

Route2Step is trained in two stages. We first initialize $\mathcal{M}_{\mathrm{IA}}$ and $\mathcal{M}_{\mathrm{AG}}$ using aligned expert trajectories. We then collect policy-induced histories through fixed-sub-instruction rollouts, using them as state-level supervision for $\mathcal{M}_{\mathrm{IA}}$ and selectively introducing expert recovery actions for $\mathcal{M}_{\mathrm{AG}}$.

\paragraph{Expert-path initialization.}
Both modules are first initialized using aligned expert trajectories. Each observation is assigned an active sub-instruction $s_t$ using the step-level alignment described above and labeled $m_t=\textsc{Normal}$. This stage establishes basic semantic-progress estimation and local sub-instruction following under expert-path observations.

\paragraph{Policy-induced rollout collection.}
After expert-path initialization, we fix the aligned active sub-instruction $s_k$ and sample multiple stochastic $\mathcal{M}_{\mathrm{AG}}$ rollouts at temperature $0.5$. A rollout succeeds only when it satisfies both the position and heading tolerances of the associated semantic waypoint. As illustrated in Fig.~\ref{fig:dagger_topdown}(c), if all sampled rollouts fail, we generate an additional intervention-assisted rollout under the same fixed semantic target.
The policy acts until its deviation from the reference segment exceeds a threshold, after which the expert guides it back near the reference trajectory and the policy resumes.
Within the intervention-assisted rollout, states before expert takeover and after reconnection are labeled $m_t=\textsc{Normal}$, whereas states during the expert-guided recovery interval are labeled $m_t=\textsc{Recovering}$. The active sub-instruction remains unchanged throughout the intervention until its semantic waypoint is reached.


\begin{table*}[!t]
\centering
\caption{
Comparison on R2R-CE and RxR-CE Val-Unseen.
Nav./VL denote additional navigation/vision--language data;
S/A denote state-/action-level policy-induced supervision.
--/n/r denote none/not reported.
Counts follow source-reported units and are not strictly comparable.
}
\label{tab:main_results}
\vspace{-1.5mm}
\scriptsize
\setlength{\tabcolsep}{3.0pt}
\renewcommand{\arraystretch}{1.05}

\resizebox{\textwidth}{!}{
\begin{tabular}{lccc|cccc|cccc}
\toprule
\multirow{2}{*}{\textbf{Method}}
& \multicolumn{2}{c}{\textbf{Extra training data}}
& \multirow{2}{*}{
  \shortstack{\textbf{Post-train}\\\textbf{supervision}}
}
& \multicolumn{4}{c|}{\textbf{R2R-CE Val-Unseen}}
& \multicolumn{4}{c}{\textbf{RxR-CE Val-Unseen}} \\
\cmidrule(lr){2-3}
\cmidrule(lr){5-8}
\cmidrule(lr){9-12}
& Nav. & VL
&
& NE$\downarrow$
& OS$\uparrow$
& SR$\uparrow$
& SPL$\uparrow$
& NE$\downarrow$
& SR$\uparrow$
& SPL$\uparrow$
& nDTW$\uparrow$ \\
\midrule

\multicolumn{12}{l}{
\textit{Panoramic RGB, odometry, and depth}
} \\




ETPNav~\cite{an2024etpnav}
& -- & --
& --
& 4.71 & 65.0 & 57.0 & 49.0
& 5.64 & 54.7 & 44.8 & 61.9 \\

HNR~\cite{wang2024lookahead}
& -- & --
& --
& 4.42 & 67.0 & 61.0 & 51.0
& 5.50 & 56.3 & 46.7 & 63.5 \\

\midrule

\multicolumn{12}{l}{
\textit{Egocentric single-view RGB}
} \\



NaVid~\cite{navid}
& -- & 763K
& 180K A
& 5.47 & 49.1 & 37.4 & 35.9
& 8.41 & 23.8 & 21.2 & -- \\

Uni-NaVid~\cite{uninavid}
& 1.2M & 2.3M
& 1.69M A
& 5.58 & 53.3 & 47.0 & 42.7
& 6.24 & 48.7 & 40.9 & -- \\

NaVILA~\cite{cheng2025navila}
& n/r & n/r
& --
& 5.22 & 62.5 & 54.0 & 49.0
& 6.77 & 49.3 & 44.0 & 58.8 \\

Aux-Think~\cite{wang2025auxthink}
& 0 & 500K
& 500K A
& 6.08 & 60.0 & 54.8 & 46.9
& 6.24 & 52.2 & 40.2 & -- \\

StreamVLN~\cite{streamvln}
& n/r & 478K
& 240K A
& 5.43 & 62.5 & 52.8 & 47.2
& 6.72 & 48.6 & 42.5 & 60.2 \\

DecoVLN~\citep{decovln}
& n/r & 178K
& 180K A
& 5.01 & 63.5 & 56.3 & 50.5 
& 5.73 & 54.2 & 46.3 & 63.5 \\


ActiveVLN~\cite{zhang2025activevln}
& 146K & --
& RL
& 5.31 & 58.3 & 50.1 & 43.7
& 5.84 & 50.7 & 41.2 & 58.1 \\


Progress-Think~\cite{wang2025progressthink}
& n/r & --
& 500K A + RL
& 4.68 & 63.6 & 60.1 & 53.6
& 8.30 & 27.5 & 22.7 & -- \\



\midrule

Route2Step (Ours)
& -- & --
& 190K S + 11.5K A
& 4.78 & 61.9 & 55.3 & 48.2
& 5.52 & 54.8 & 42.6 & 58.0 \\

\bottomrule
\end{tabular}
}
\vspace{-1.5mm}
\end{table*}

\paragraph{State-level supervision.}

For each policy-induced history, the active sub-instruction is preserved until its associated semantic waypoint is physically reached. Because completion is evaluated in physical space rather than by the temporal index of the expert trajectory, the same semantic boundary remains applicable when the agent deviates, loops, or reconnects to the route. This geometry-grounded relabeling converts off-path visual histories into physically grounded semantic-progress supervision for $\mathcal{M}_{\mathrm{IA}}$. The execution-state label $m_t\in\{\textsc{Normal},\textsc{Recovering}\}$ indicates whether the current semantic step should be followed nominally or through corrective execution. These samples supervise the semantic state without prescribing the next low-level action.

\paragraph{Selective action-level correction.}

States labeled $m_t=\textsc{Recovering}$ in the intervention-assisted rollout retain the same semantic target $s_t$ and are paired with the corresponding expert recovery action chunks to supervise $\mathcal{M}_{\mathrm{AG}}$.
These samples expose $\mathcal{M}_{\mathrm{AG}}$ to off-path observations together with expert-guided recovery behavior.
In contrast, states labeled $m_t=\textsc{Normal}$ in the same rollout provide state-level supervision without additional expert-action targets.
This concentrates expert-action supervision on persistent local-execution failures of $\mathcal{M}_{\mathrm{AG}}$ under the correct active sub-instruction.

\paragraph{Training objectives.}
Let $\mathcal{D}_{E}$ denote expert-path samples, $\mathcal{D}_{S}$ state-level samples collected from fixed-sub-instruction rollouts, and $\mathcal{D}_{A}$ expert-recovery samples from the $\textsc{Recovering}$ intervals of unsuccessful rollout groups. All rollout histories contribute semantic-state targets to $\mathcal{D}_{S}$, while only the $\textsc{Recovering}$ interval of an unsuccessful group additionally contributes expert-action targets to $\mathcal{D}_{A}$.

\begin{align}
\mathcal{L}_{\mathrm{IA}}
&=
-\mathbb{E}_{\mathcal{D}_E\cup\mathcal{D}_S}
\log p_{\theta_{\mathrm{IA}}}
\!\left(
s_t,m_t\mid\mathcal{I},V_{1:t}
\right),
\\
\mathcal{L}_{\mathrm{AG}}
&=
-\mathbb{E}_{\mathcal{D}_E\cup\mathcal{D}_A}
\nonumber\\[-2pt]
&\quad
\log p_{\theta_{\mathrm{AG}}}
\!\left(
a_{t:t+h}\mid
\mathcal{I},s_t,m_t,V_{t-k:t}
\right).
\end{align}

No gradient is propagated across the semantic--execution interface. This formulation allows policy-induced histories to improve semantic-state prediction without uniformly converting them into action labels, while direct action supervision is concentrated on recovery states collected after repeated fixed-sub-instruction failures.
\section{Experiments}
\label{sec:experiments}

\subsection{Experimental Setup}
\label{sec:exp_setup}

\noindent\textbf{Implementation and Training Details.}
Route2Step consists of two independently trained Qwen2.5-VL-3B models~\citep{bai2025qwen2_5vl}: the Instruction Analysis Module $\mathcal{M}{\mathrm{IA}}$ and the Action Generation Module $\mathcal{M}{\mathrm{AG}}$. Both modules are optimized with supervised fine-tuning using learning rates of $1\times10^{-5}$ for expert-path initialization and $5\times10^{-6}$ for corrective training. All experiments are conducted on 4 NVIDIA PRO 6000 GPUs.

\noindent\textbf{Evaluation.}
We evaluate on the R2R-CE and RxR-CE Val-Unseen splits, both built on Matterport3D~\citep{chang2017matterport3d}, following the public StreamVLN sensor--action protocol: egocentric RGB observations at $640\times480$ resolution with a $79^\circ$ horizontal field of view, 0.25\,m forward steps, $15^\circ$ turns, a 500-step limit, and a 3\,m success radius. We report navigation error (NE), oracle success (OS), success rate (SR), and success weighted by path length (SPL) on both benchmarks, together with normalized dynamic time warping (nDTW) on RxR-CE~\citep{ilharco2019ndtw}. We use SR and SPL as the primary metrics.

\begin{table}[t]
\centering
\caption{
Ablation of corrective-supervision allocation on R2R-CE
Val-Unseen. All variants share the same expert-path
initialization. State and Action denote the numbers of
state-level and directly action-supervised corrective samples.
}
\label{tab:correction_allocation}
\small
\renewcommand{\arraystretch}{1.15}
\begin{tabular*}{\columnwidth}{
@{\extracolsep{\fill}}lcccc@{}
}
\toprule
Strategy & State & Action & SR$\uparrow$ & SPL$\uparrow$ \\
\midrule
Expert-path only
& -- & -- & 48.1 & 43.3 \\

State-level correction
& 190K & -- & 50.6 & 44.7 \\

Selective action
& -- & 11.5K & 49.4 & 44.8 \\

Conventional DAgger
& -- & 200K & 49.8 & 44.4 \\

State + conventional
& 190K & 200K & 54.4 & 46.7 \\

Level-separated correction
& 190K & 11.5K & \textbf{55.3} & \textbf{48.2} \\
\bottomrule
\end{tabular*}
\end{table}
\subsection{Main Navigation Performance}

As shown in Table~\ref{tab:main_results}, Route2Step achieves 55.3/48.2 SR/SPL on R2R-CE and 54.8/42.6 on RxR-CE, establishing competitive performance without extra navigation or general vision--language data. 
The consistent results across both benchmarks indicate that the benefit of explicit semantic-progress modeling extends beyond a single dataset or instruction distribution.

\subsection{Analysis of Level Separation}

\noindent\textbf{State- versus action-level correction.}
Table~\ref{tab:correction_allocation} examines supervision allocation
across the two temporal levels. Action-level correction alone yields limited gains: selective correction with 11.5K labels reaches 49.4 SR and 44.8 SPL, while conventional DAgger with 200K labels reaches 49.8 SR and 44.4 SPL. Because $\mathcal{M}_{\mathrm{AG}}$ operates only on recent observations, these local action targets cannot substitute for correcting route-level progress from the accumulated history.

Adding 190K state-level samples to $\mathcal{M}_{\mathrm{IA}}$ and only
11.5K selective action labels raises performance to 55.3 SR and 48.2 SPL. With the same state-level correction, selective action supervision outperforms 200K conventional action labels by 0.9 SR and 1.5 SPL points. These results support correcting semantic progress from accumulated visual history while reserving action supervision for persistent local-execution failures.

\noindent\textbf{Role of the execution-state interface.}
We examine whether the gains arise solely from recovery-oriented data or also from explicitly conditioning $\mathcal{M}_{\mathrm{AG}}$ on the execution state. The implicit-recovery variant uses recovery-oriented observations without conditioning $\mathcal{M}_{\mathrm{AG}}$ on $m_t$, and its improvement over expert-path training shows that such observations are beneficial even without explicit state conditioning. The last two variants share the same state-conditioned $\mathcal{M}_{\mathrm{AG}}$ checkpoint and differ only in whether $m_t$ is fixed or predicted at inference. Using the predicted $m_t$ improves SR from 53.2 to 55.3 and SPL from 47.6 to 48.2, showing that the execution state is an operational component of the semantic--execution interface, routing the same semantic target to nominal or corrective local behavior.

\begin{table}[t]
\centering
\caption{
Effect of the execution-state interface on R2R-CE Val-Unseen.
The same $\mathcal{M}_{\mathrm{IA}}$ is used in all settings. The last two rows share the same
state-conditioned $\mathcal{M}_{\mathrm{AG}}$ checkpoint and differ only in test-time routing.
}
\label{tab:execution_state}
\small
\setlength{\tabcolsep}{5pt}
\renewcommand{\arraystretch}{1.05}
\begin{tabular}{lccc}
\toprule
Setting
& State-cond. $\mathcal{M}_{\mathrm{AG}}$
& SR$\uparrow$
& SPL$\uparrow$ \\
\midrule
Implicit recovery
& \xmark
& 51.0
& 44.1 \\

Fixed \textsc{Normal}
& \cmark
& 53.2
& 47.6 \\

Predicted state
& \cmark
& \textbf{55.3}
& \textbf{48.2} \\
\bottomrule
\end{tabular}
\end{table}

\noindent\textbf{Semantic-progress robustness.}
To isolate semantic-progress tracking from local action execution, we directly evaluate $\mathcal{M}_{\mathrm{IA}}$ on the human-aligned FG-R2R and Landmark-RxR Val-Unseen splits.
The On-path setting uses histories from the original expert trajectories, while the Deviated setting applies four model-independent perturbations: heading deviation, lateral detour, short backtracking, and local loop.

Because $\mathcal{M}_{\mathrm{IA}}$ generates free-form text, each prediction is matched to the most similar human-annotated sub-instruction from the same route. Strict active-sub-instruction accuracy counts a query as correct only when the matched sub-instruction is the human-aligned active one. The reported values are query-level accuracies rather than semantic-similarity scores.

As shown in Table~\ref{tab:mia_direct}, expert-path training and state-level correction yield nearly identical accuracy on both On-path splits. In contrast, state-level correction improves strict active-sub-instruction accuracy by 10.51 points on FG-R2R and 8.26 points on Landmark-RxR under Deviated histories. State-level correction therefore improves semantic-progress robustness under spatial deviations, with little effect on on-path accuracy.
\begin{table}[t]
  \centering
    \caption{
    Strict active-sub-instruction accuracy on human-aligned Val-Unseen
    splits under On-path and Deviated histories.
    All values are percentages.
    }
  \label{tab:mia_direct}
  \small
  \setlength{\tabcolsep}{4.2pt}
  \renewcommand{\arraystretch}{1.08}

  \begin{tabular*}{\columnwidth}{
    @{\extracolsep{\fill}}lcccc@{}
  }
  \toprule
  \multirow{2}{*}{Setting}
  & \multicolumn{2}{c}{FG-R2R}
  & \multicolumn{2}{c}{Landmark-RxR} \\
  \cmidrule(lr){2-3}
  \cmidrule(lr){4-5}
  & On-path & Deviated & On-path & Deviated \\
  \midrule

  $\mathcal{M}_{\mathrm{IA}}$ 3B, expert
  & 58.93
  & 43.54
  & 70.33
  & 56.13 \\

  $\mathcal{M}_{\mathrm{IA}}$ 3B, corrected
  & 58.61
  & 54.05
  & 69.99
  & 64.39 \\
  
  \midrule

  Single 7B, corrected
  & 60.66
  & 55.55
  & 71.62
  & 65.68 \\

  \bottomrule
  \end{tabular*}
\end{table}

\noindent\textbf{Comparison with a unified policy.}
To test whether a larger unified policy can benefit from the same semantic-progress supervision without an explicit inference-time interface, we train a single-system 7B baseline using the complete expert-path and corrective training data available to Route2Step.
The model jointly learns active-sub-instruction prediction and action generation during training, but directly predicts actions at inference without exposing semantic progress through an explicit semantic--execution interface. On R2R-CE Val-Unseen, this baseline achieves 50.7\% SR and 44.1\% SPL, compared with 55.3\% SR and 48.2\% SPL for Route2Step.

Using the same human-aligned protocol, the 7B baseline slightly outperforms the 3B $\mathcal{M}_{\mathrm{IA}}$ trained with state-level correction on both On-path and Deviated histories (Table~\ref{tab:mia_direct}). Nevertheless, its navigation performance remains 4.6 SR and 4.1 SPL points below Route2Step. This contrast indicates that a unified model can learn an accurate semantic-progress representation, but auxiliary progress supervision alone does not translate this signal into navigation gains as effectively as an explicit interface that separates semantic progress from local execution.

\begin{figure*}[t]
    \centering
    \makebox[\textwidth][c]{%
        \includegraphics[width=0.96\textwidth]
        {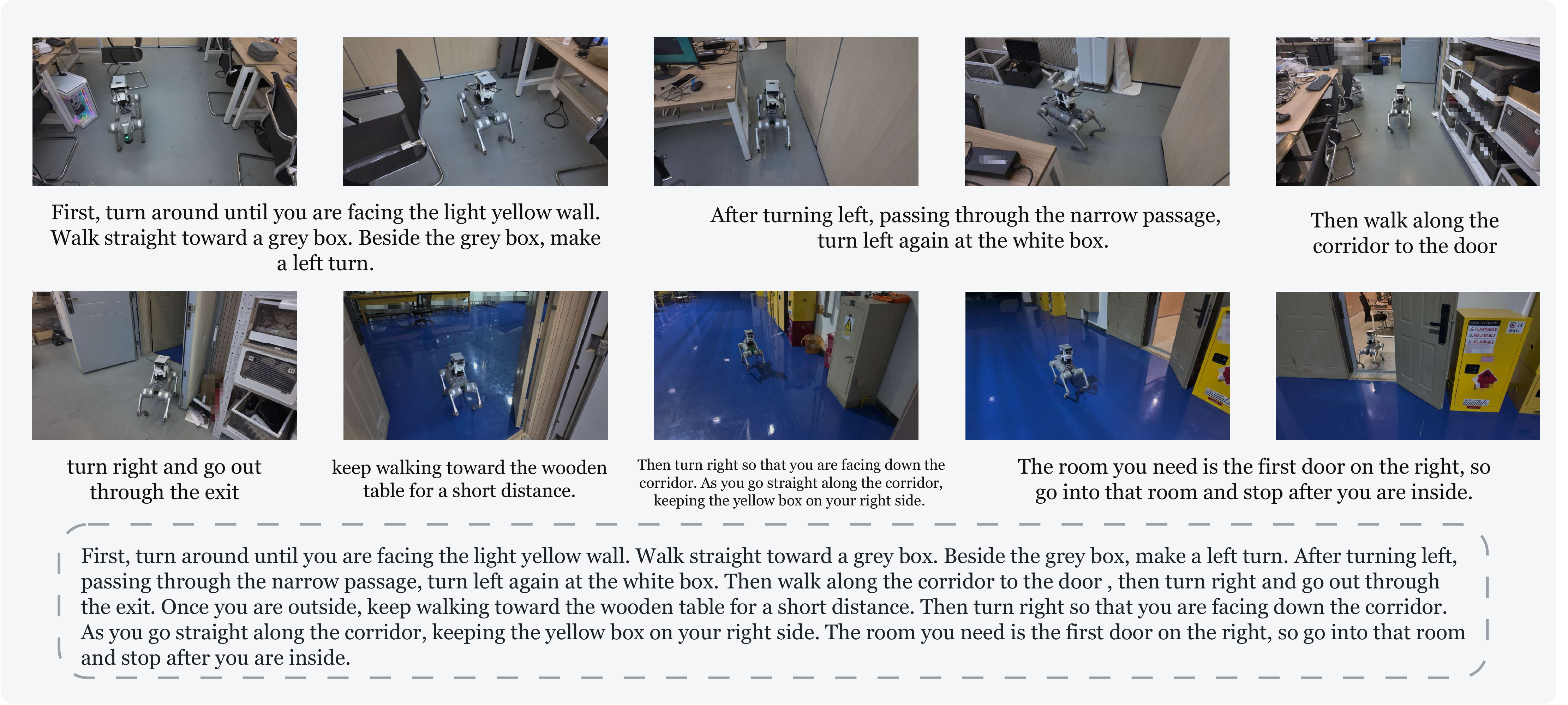}%
    } 
    \caption{
    A real-world execution trace in a complex indoor environment.
    Observations are annotated with the active sub-instruction predicted
    by $\mathcal{M}_{\mathrm{IA}}$, providing an interpretable account of
    semantic progress during long-horizon navigation.
    }
    \label{fig:realworld}
\end{figure*}

\noindent\textbf{Transferability of the semantic interface.}
\begin{table}[bp]
  \centering
  \caption{
Cross-policy transfer on R2R-CE Val-Unseen. Frozen policies are augmented with the predicted active sub-instruction without additional training.
  }
  \label{tab:plugin_transfer}
  \small
  \setlength{\tabcolsep}{4.5pt}
  \renewcommand{\arraystretch}{1.08}
  \begin{tabular*}{\columnwidth}{
    @{\extracolsep{\fill}}lcccc@{}
  }
  \toprule
  \multirow{2}{*}{Policy}
  & \multicolumn{2}{c}{Base}
  & \multicolumn{2}{c}{$+\mathcal{M}_{\mathrm{IA}}$} \\
  \cmidrule(lr){2-3}
  \cmidrule(lr){4-5}
  & SR$\uparrow$ & SPL$\uparrow$
  & SR$\uparrow$ & SPL$\uparrow$ \\
  \midrule
  NaVid
  & 42.5
  & 37.5
  & \textbf{47.4}
  & \textbf{41.4} \\

  Uni-NaVid
  & 52.1
  & 47.8
  & \textbf{54.6}
  & \textbf{49.7} \\

  NaVILA
  & 54.3
  & 49.6
  & \textbf{55.4}
  & \textbf{51.1} \\

  StreamVLN
  & 56.8
  & 50.5
  & \textbf{57.4}
  & \textbf{52.3} \\
  \bottomrule
  \end{tabular*}
\end{table}
To test whether the active sub-instruction predicted by $\mathcal{M}_{\mathrm{IA}}$ is useful beyond $\mathcal{M}_{\mathrm{AG}}$, we inject it into four frozen VLN policies: NaVid, Uni-NaVid, NaVILA, and StreamVLN. At each decision step, the predicted active sub-instruction is appended to the textual input of the corresponding policy.
For each policy, the Base and $+\mathcal{M}_{\mathrm{IA}}$ variants use the same checkpoint, inference configuration, and local evaluation environment, differing only in the injected active sub-instruction; no additional training or parameter updates are performed.

As shown in Table~\ref{tab:plugin_transfer}, providing the predicted active sub-instruction improves both SR and SPL for all four frozen policies without additional training. The improvements persist across substantially different architectures and baseline performance, indicating that the benefit is not tied to a specific execution policy or performance regime. Making semantic progress explicit at inference therefore provides actionable guidance beyond the progress information these policies infer implicitly from their original inputs. The active sub-instruction serves as a transferable interface for local execution, rather than merely a representation tailored to $\mathcal{M}_{\mathrm{AG}}$.

\subsection{Real-World Evaluation}

We deploy Route2Step on a Unitree GO2 equipped with an Intel RealSense D455 camera, using only its single-view RGB stream. The robot continuously transmits egocentric observations to a remote workstation equipped with an NVIDIA PRO 6000 GPU, where $\mathcal{M}_{\mathrm{IA}}$ and $\mathcal{M}_{\mathrm{AG}}$ perform inference and return executable action chunks for closed-loop control. The model is transferred directly from simulation without real-world fine-tuning and operates without depth, localization, mapping, or a pre-built environment model.

\noindent\textbf{Deployment across diverse environments.}
Across nine tasks in laboratories, a cafe, residential areas, parks, and a parking area, Route2Step completes 19 of 33 autonomous trials, including 5/10 trials on two complex indoor tasks and 14/23 trials on seven outdoor tasks. The instructions range from 14 to 120 words and cover narrow passages, repeated landmarks, and visually ambiguous locations. 
Detailed task descriptions, per-task results, and representative execution records are provided in the supplementary material. Fig.~\ref{fig:realworld} shows a representative long-horizon execution.

\noindent\textbf{Indoor comparison on a challenging task.}
We further compare StreamVLN, StreamVLN $+\mathcal{M}_{\mathrm{IA}}$, and Route2Step on the most challenging 120-word indoor task, with five trials per method under matched task conditions. We report full-task success and the average number of completed sub-instructions, counting a step as completed when its corresponding semantic waypoint is reached.

StreamVLN fails to complete the task in all trials and reaches an average of 2.2 out of the seven semantic waypoints. Providing the predicted active sub-instruction raises StreamVLN's average progress from 2.2 to 4.0 waypoints, although both variants remain at 0/5 success. Route2Step reaches an average of 5.0 waypoints and succeeds in three out of five trials. This comparison suggests that explicit semantic-progress guidance improves intermediate task progress, while semantic guidance and interface-conditioned local execution play complementary roles in completing the full route.

\begin{table}[t]
\centering
\caption{Indoor comparison on the complex task under matched task
conditions.}
\label{tab:realworld_controlled}
\small
\setlength{\tabcolsep}{4.2pt}
\renewcommand{\arraystretch}{1.08}
\begin{tabular*}{\columnwidth}{
  @{\extracolsep{\fill}}lccc@{}
}
\toprule
Metric
& StreamVLN
& \shortstack{StreamVLN $+\mathcal{M}_{\mathrm{IA}}$}
& Route2Step \\
\midrule
Success
& 0 / 5
& 0 / 5
& \textbf{3 / 5} \\

Avg. progress
& 2.2 / 7
& 4.0 / 7
& \textbf{5.0 / 7} \\
\bottomrule
\end{tabular*}
\end{table}
\section{Conclusion}

We presented Route2Step, a level-separated VLN framework that separates semantic-progress tracking from local execution through an explicit active-sub-instruction and execution-state interface. E-SPA provides the required step-level supervision by aligning sub-instructions with trajectory segments and pose-level completion targets. This decomposition further enables corrective supervision to be assigned by temporal scope: policy-induced histories provide broad state-level correction for semantic progress, whereas direct expert actions are reserved for persistent local-execution failures.
On R2R-CE, Route2Step improves SR from 48.1\% to 55.3\% and SPL from 43.3\% to 48.2\%, using 190K state-level corrective samples but only 11.5K directly action-supervised states, and remains effective on RxR-CE. Ablations, direct progress evaluation, and cross-policy transfer consistently support the benefit of exposing semantic progress as an explicit interface for execution. Real-world deployments further demonstrate its practical applicability beyond simulation.

\bibliography{references}

\clearpage

\appendix

\section{Alignment and Supervision Construction}

\subsection{E-SPA Alignment Procedure}
\label{app:espa}

\subsubsection{Inputs, step normalization, and feasible segments.}

For each R2R demonstration, E-SPA uses the route instruction $\mathcal I$, the complete RGB observation sequence $\mathcal V$, the associated low-level action sequence $\mathcal A$, and the Habitat pose trajectory $\mathcal P$. The instruction-normalization stage obtains an ordered sequence of locally executable sub-instructions $S=(s_1,\ldots,s_n)$ from $\mathcal I$. The visual and action sequences, together with the height profile derived from $\mathcal P$, are subsequently used for trajectory alignment, while the full endpoint poses provide the position and heading of the resulting semantic waypoints. The numerical settings used by E-SPA are summarized in Table~\ref{tab:app_espa_config}.

Each retained step is intended to contain an executable navigation action or goal. Purely descriptive clauses are not introduced as standalone steps, although landmark and scene descriptions may remain attached to the corresponding navigation action. Split or merged steps preserve the source ordering.

Let $T$ denote the number of trajectory frames used for alignment, and let $\mathcal F_k$ denote the feasible frame intervals for sub-instruction $s_k$. The final alignment consists of non-empty, ordered, and contiguous segments that cover the complete trajectory:
\begin{equation}
\begin{aligned}
\mathcal B=\bigl\{B:\; &b_1=1<\cdots<b_{n+1}=T+1,\\[-2pt]
&[b_k,b_{k+1}-1]\in\mathcal F_k\bigr\}.
\end{aligned}
\end{equation}

E-SPA constructs the R2R training alignments without using FG-R2R annotations or other human-provided temporal sub-instruction alignments. FG-R2R is reserved exclusively for evaluation.

\subsubsection{Cost definitions and dynamic program.}
\label{app:espa_cost}

\paragraph{Semantic-alignment cost.}
Let $\phi_v(V_t)$ and $\phi_s(s_k)$ denote the visual and textual features extracted by the frozen CLIP encoder. We first apply $\ell_2$ normalization:
\begin{equation}
\bar{v}_t =
\frac{\phi_v(V_t)}{\|\phi_v(V_t)\|_2},
\qquad
\bar{s}_k =
\frac{\phi_s(s_k)}{\|\phi_s(s_k)\|_2}.
\end{equation}
For each frame $t$, the semantic cost is the negative log-probability of sub-instruction $s_k$ under a temperature-scaled softmax over all $n$ sub-instructions from the same route:
\begin{equation}
c_{\rm sem}(t,k)
=
-\log
\frac{
\exp\!\left(\bar{v}_t^{\top}\bar{s}_k/\tau\right)
}{
\sum_{\ell=1}^{n}
\exp\!\left(\bar{v}_t^{\top}\bar{s}_{\ell}/\tau\right)
},
\qquad \tau=0.05.
\label{eq:app_frame_sem_cost}
\end{equation}
The semantic cost of assigning $s_k$ to a candidate segment $[i,j]$ is
obtained by summing the frame-level costs:
\begin{equation}
C_{\rm sem}(s_k,i,j)
=
\sum_{t=i}^{j} c_{\rm sem}(t,k).
\label{eq:app_sem_cost}
\end{equation}


\begin{table}[t]
\centering
\caption{E-SPA implementation settings.}
\label{tab:app_espa_config}
\scriptsize
\setlength{\tabcolsep}{4pt}
\renewcommand{\arraystretch}{1.08}
\begin{tabular}{lc}
\toprule
Setting & Value \\
\midrule
Step-normalization model
& \texttt{qwen3-vl-flash} \\
Step-normalization temperature
& $0$ \\
Stair-association model
& \texttt{qwen3-vl-plus} \\
Stair-association temperature
& $0.1$ \\
Visual encoder
& CLIP ViT-L/14, $224\times224$ \\
Semantic temperature $\tau$
& $0.05$ \\
Action intents
& Forward, Left, Right, Stop \\
Maximum candidate length
& $2T/n$ \\
Per-step / span height thresholds 
& 0.01 m / 0.08 m \\
$\lambda_{\rm sem}$
& $0.4$ \\
$\lambda_{\rm act}$
& $0.6$ \\
$\lambda_{\rm dur}$
& $0.05$ \\
\bottomrule
\end{tabular}
\end{table}

\paragraph{Action-consistency cost.}
Each normalized sub-instruction is assigned an offline motion-intent label
from
\{\textsc{Forward}, \textsc{Left}, \textsc{Right}, \textsc{Stop}\}
using a Qwen-based mapping procedure. The resulting mapping is fixed during
trajectory alignment. Most sub-instructions have a single intent, while
explicitly direction-ambiguous turns, such as turning around, may admit both
\textsc{Left} and \textsc{Right}.

The observed primitive action at frame $t$ is represented by a
three-dimensional motion vector $\mathbf m_t$:
\begin{equation}
\begin{aligned}
\mathbf e_{\rm stop} &= [0,0,0], &
\mathbf e_{\rm forward} &= [0,0,-1],\\
\mathbf e_{\rm left} &= [0,1,0], &
\mathbf e_{\rm right} &= [0,-1,0].
\end{aligned}
\label{eq:app_action_vectors}
\end{equation}
The initialization frame uses the same zero vector as \textsc{Stop}.
Let $\mathcal Y_k$ denote the set of valid intent vectors for
sub-instruction $s_k$. The frame-level action-consistency cost is
\begin{equation}
c_{\rm act}(t,k)
=
\frac{1}{2}
\min_{\mathbf y\in\mathcal Y_k}
\left\|\mathbf m_t-\mathbf y\right\|_2^2.
\label{eq:app_frame_act_cost}
\end{equation}
The action cost of assigning $s_k$ to a candidate segment $[i,j]$ is
obtained by summing the frame-level costs:
\begin{equation}
C_{\rm act}(s_k,i,j)
=
\sum_{t=i}^{j} c_{\rm act}(t,k).
\label{eq:app_act_cost}
\end{equation}
Stair-related geometry is handled separately through the stair-partition constraints described below; it does not alter the action representation.

\paragraph{Duration regularization.}
To discourage highly imbalanced partitions, E-SPA penalizes deviations from
the uniform target segment length. Let
\[
\bar{L}=\frac{T}{n}
\]
denote the ideal number of frames assigned to each sub-instruction. For a
candidate segment $[i,j]$ with length $L_{i:j}=j-i+1$, the duration cost is
\begin{equation}
C_{\rm dur}(i,j)
=
\left(
L_{i:j}-\frac{T}{n}
\right)^2.
\label{eq:app_dur_cost}
\end{equation}
This cost is weighted by $\lambda_{\rm dur}=0.05$ and added once for each candidate segment during the dynamic-program transition rather than being included in the frame-level cost matrix.

\paragraph{Stair anchoring.}

For stair-related demonstrations, E-SPA introduces a geometry-based stair anchor implemented as a hard constraint. Candidate vertical-motion spans are detected from the height profile derived from Habitat poses. Consecutive states with an absolute height change of at least $0.01$\,m are grouped into candidate spans, which are retained only when their endpoint height difference is at least $0.08$\,m. A text-only selector then associates ordered stair-related sub-instructions with the retained spans.

When a valid association is obtained, the trajectory and sub-instruction sequence are partitioned into pre-stair, stair, and post-stair blocks. Dynamic programming is performed independently within each block, so boundaries cannot cross the selected stair-block boundaries, while internal boundaries remain optimized by the same procedure.

Let $\mathcal F_k^{\rm stair}$ denote the feasible segment intervals after applying the stair constraint. The anchor term is implemented as a binary-gated hard constraint:
\begin{equation}
C_{\rm anchor}(s_k,i,j)
=
\begin{cases}
0, & [i,j]\in\mathcal F_k^{\rm stair},\\
+\infty, & [i,j]\notin\mathcal F_k^{\rm stair}.
\end{cases}
\end{equation}
For non-stair demonstrations,
$\lambda_{\mathrm{anchor}}=0$ and the anchor term is inactive.
When a valid stair anchor is identified, $\lambda_{\mathrm{anchor}}=1$; infeasible intervals crossing the selected stair boundaries are assigned infinite cost and excluded from the dynamic-programming search space.

\paragraph{Combined objective and dynamic program.}

The complete alignment objective consists of semantic, action, and duration
costs, while the stair anchor is incorporated as a hard constraint on the feasible transition space. The semantic and action terms form the frame-level cost:
\begin{equation}
L_{t,k}
=
\lambda_{\rm sem}c_{\rm sem}(t,k)
+
\lambda_{\rm act}c_{\rm act}(t,k),
\label{eq:app_frame_cost}
\end{equation}
where $\lambda_{\rm sem}=0.4$ and $\lambda_{\rm act}=0.6$.
The duration term is applied on each DP transition.

The hard constraint induced by stair anchoring modifies the feasible transition space rather than introducing an additional finite cost. Therefore, the same dynamic programming formulation is used for both stair and non-stair demonstrations,
while infeasible assignments crossing the selected stair boundaries are
removed from the candidate set.

Let $D[k,t]$ denote the minimum cost of assigning frames
$1,\ldots,t$ to sub-instructions $s_1,\ldots,s_k$. The dynamic program is
\begin{equation}
\begin{aligned}
D[1,t]
&=
\sum_{r=1}^{t} L_{r,1}
+
\lambda_{\rm dur}C_{\rm dur}(1,t),\\
D[k,t]
&=
\min_{p\in\mathcal P(k,t)}
\Bigg\{
D[k-1,p]
+
\sum_{r=p+1}^{t}L_{r,k}\\
&\hspace{4.2em}
+
\lambda_{\rm dur}C_{\rm dur}(p+1,t)
\Bigg\},
\qquad k\ge2.
\end{aligned}
\label{eq:app_dp}
\end{equation}
where $\mathcal P(k,t)$ enforces non-empty segments, the
$2T/n$ candidate-length constraint, and the stair-anchor constraints
for stair-related demonstrations.

Each segment contains at least one frame. Candidate segments longer than $2T/n$ are pruned. Backtracking from $D[n,T]$ yields the ordered boundaries $1=b_1<\cdots<b_{n+1}=T+1$. The endpoint pose of segment $[b_k,b_{k+1}-1]$ is retained as the semantic waypoint $w_k=(p_k,\theta_k)$, including both position and heading.



\subsection{Dataset Mapping and Corrective Supervision}
\label{app:data}

\subsubsection{Mapping human annotations to VLN-CE trajectories.}
\label{app:mapping}


For Landmark-RxR and FG-R2R, the annotated sub-path endpoints are first located on the corresponding discrete reference path and then monotonically projected onto the continuous VLN-CE trajectory. Discrete viewpoint indices are not treated as continuous frame indices; instead, each reference pose is matched to the nearest continuous state under Euclidean position distance while preserving the annotation order. The final segment is extended to the last trajectory frame to ensure complete coverage.

For R2R-CE, the continuous trajectory is reconstructed from the official action and location records. For RxR-CE, we use trajectories executed with the same $15^\circ$ action space as our evaluation protocol rather than the original $30^\circ$ guide-action sequence.

\paragraph{Turn-boundary refinement.}
Because in-place rotations produce multiple consecutive frames at nearly identical positions, pose-based projection may assign a semantic boundary to the wrong side of a rotation sequence. We therefore detect local position-invariant rotation spans around each projected boundary and reassign them according to whether the adjacent sub-instructions semantically require the turn. If both adjacent steps require turning, the span is divided between them; otherwise, the original boundary is retained when no reassignment is supported.

\subsubsection{Geometry-grounded corrective rollouts.}
\label{app:corrective_rollouts}

\paragraph{Waypoint completion and rollout collection.}
For each aligned sub-instruction $s_k$, let
$w_k=(p_k,\theta_k)$ denote the endpoint pose of its expert segment.
During policy rollout, an intermediate waypoint is considered reached when
\begin{equation}
\|p_t-p_k\|_2 \le 1.5~\mathrm{m},
\qquad
d_{\angle}(\theta_t,\theta_k)\le45^\circ,
\label{eq:app_waypoint_reached}
\end{equation}
where $d_{\angle}$ denotes the wrapped angular distance. For the final route
goal, only the positional criterion $\|p_t-p_k\|_2\le2.0~\mathrm{m}$ is used.

For each fixed sub-instruction, MAG performs multiple stochastic rollouts from the same segment-start state with temperature 0.5. Each rollout is bounded by a horizon proportional to the corresponding expert-segment length. A rollout terminates when $\mathcal{M}_{\mathrm{AG}}$ predicts \textsc{Stop} or reaches the horizon. A rollout is considered successful once the associated semantic waypoint is reached.

\paragraph{Intervention-based recovery.}

\begin{table}[t]
\centering
\caption{Deviation and reconnection thresholds used during corrective
rollout collection.}
\label{tab:app_recovery_thresholds}
\small
\setlength{\tabcolsep}{5pt}
\begin{tabular}{lcc}
\toprule
Expert-segment length & Takeover & Reconnection \\
\midrule
$L_k^{\rm GT}>30$ & $2.0$ m & $0.8$ m \\
$10<L_k^{\rm GT}\le30$ & $1.0$ m & $0.5$ m \\
$L_k^{\rm GT}\le10$ or unavailable & $0.5$ m & $0.3$ m \\
\bottomrule
\end{tabular}
\end{table}

If all ordinary rollouts fail to complete the current subgoal, we collect one
expert-intervention trajectory while keeping $s_k$ fixed. During expert
intervention, the current subgoal is considered completed only when the agent
reaches the endpoint pose of the corresponding expert segment:
\begin{equation}
\|p_t-p_k\|_2 \le 0.5~\mathrm{m},
\qquad
d_{\angle}(\theta_t,\theta_k)\le15^\circ.
\label{eq:app_precise_waypoint}
\end{equation}
This stricter criterion is used to terminate the expert-intervention rollout.

In the reported experiments, policy deviation is measured as the minimum
three-dimensional Euclidean distance to the reference geometry of the current
expert segment:
\[
d_{\rm dev}(p_t)
=
\min_{q\in\mathcal P_k^{GT}}
\|p_t-q\|_2 .
\]
Here, $\mathcal P_k^{GT}$ denotes the reference trajectory of the current
segment.

The takeover and reconnection thresholds depend on the expert-segment length.
For recoverable interventions, control is returned to the policy once the
deviation falls below the corresponding reconnection threshold and the
heading error relative to the segment endpoint is at most $15^\circ$. This heading condition is used as a conservative handoff heuristic during
corrective-rollout collection and favors reliable reconnection over early
policy release.

\paragraph{Supervision allocation.}

The collected corrective trajectories are used differently by the two modules. All retained policy-induced histories, including ordinary and intervention-assisted rollouts, provide active-sub-instruction supervision for $\mathcal{M}_{\mathrm{IA}}$. Action-level supervision is retained only from the expert-guided recovery intervals.

\section{Model Implementation and Training}

\paragraph{Temporal context and interface.}
$\mathcal{M}_{\mathrm{IA}}$ receives at most 16
chronologically ordered observations: up to 13 historical
views sampled using a power-law schedule with exponent 1.5,
followed by the three most recent consecutive views.
$\mathcal{M}_{\mathrm{AG}}$ receives at most eight views
sampled with the same schedule from the latest 40
observations and predicts up to three primitive actions per
call. The active sub-instruction and execution state are
jointly serialized in the textual interface passed from
$\mathcal{M}_{\mathrm{IA}}$ to $\mathcal{M}_{\mathrm{AG}}$,
with \texttt{Recovering:} used to indicate corrective
execution.

\paragraph{Training configuration.}
Both modules are optimized using supervised fine-tuning.
The vision transformer is frozen, while the visual merger
and language-side parameters remain trainable. The complete
training configuration is summarized in
Table~\ref{tab:app_training_config}.

\begin{table}[t]
\centering
\caption{Training configuration}
\label{tab:app_training_config}
\scriptsize
\setlength{\tabcolsep}{3.5pt}
\renewcommand{\arraystretch}{1.08}
\begin{tabular}{lcc}
\toprule
Configuration & $\mathcal M_{\rm IA}$ & $\mathcal M_{\rm AG}$ \\
\midrule
Base model & Qwen2.5-VL-3B & Qwen2.5-VL-3B \\
Vision transformer & Frozen & Frozen \\
Number of GPUs & 4 & 4 \\
Global batch size & 128 & 128 \\
Expert-path initialization learning rate
& $1\times10^{-5}$ & $1\times10^{-5}$ \\
Subsequent learning rate
& $5\times10^{-6}$ & $5\times10^{-6}$ \\
Optimizer & AdamW fused & AdamW fused \\
Scheduler & Cosine & Cosine \\
Warmup ratio & 0.03 & 0.03 \\
Weight decay & 0.1 & 0.1 \\
Gradient clipping & 1.0 & 1.0 \\
Precision & BF16 & BF16 \\
\bottomrule
\end{tabular}
\end{table}
\begin{figure*}[t]
    \centering
    \setlength{\tabcolsep}{3pt}

    \begin{tabular}{@{}cccc@{}}
        \includegraphics[width=0.15\textwidth]{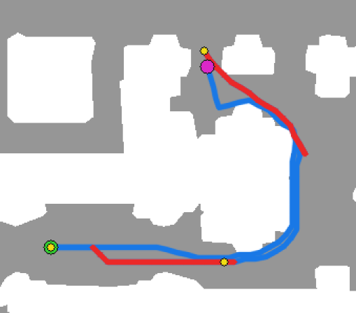}
        &
        \includegraphics[width=0.15\textwidth]{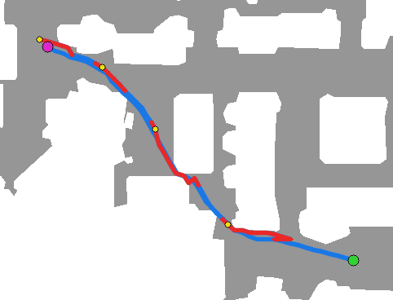}
        &
        \includegraphics[width=0.2\textwidth]{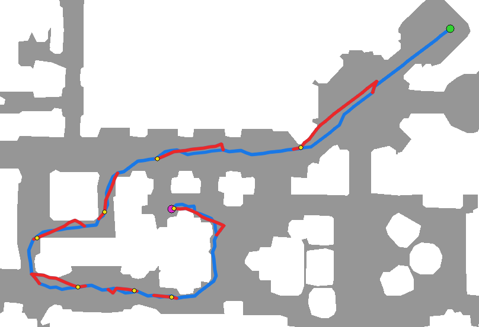}
        &
        \includegraphics[width=0.2\textwidth]{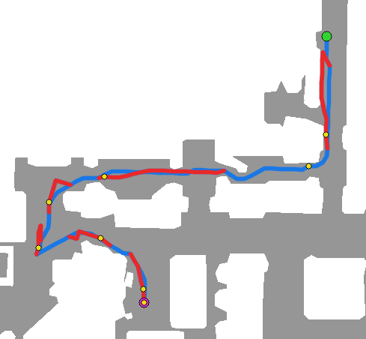}
        \\[-1pt]

        {\scriptsize (a)}
        &
        {\scriptsize (b)}
        &
        {\scriptsize (c)}
        &
        {\scriptsize (d)}
    \end{tabular}

    \caption{
    Representative top-down visualizations of controlled
    off-path histories used for direct evaluation of
    $\mathcal{M}_{\mathrm{IA}}$. Blue denotes the on-path
    trajectory, while red denotes the constructed deviation
    and reconnection trajectory. The ground-truth active
    sub-instruction remains fixed throughout each sequence.
    }
    \label{fig:mia_direct}
\end{figure*}

\paragraph{Training data and supervision accounting.}
Expert-path initialization uses approximately 200K $\mathcal{M}_{\mathrm{IA}}$ samples, comprising 100K from R2R and 100K from RxR, and 650K $\mathcal{M}_{\mathrm{AG}}$ action samples, comprising 225K from R2R and 425K from RxR. The same 3B expert-path initialization is used across the
corrective-supervision ablations, with a learning rate of $1\times10^{-5}$. Subsequent training mixes expert-path and corrective samples using a learning rate of $5\times10^{-6}$.

The corrective-supervision budgets reported in the main paper exclude the expert-path initialization and count only the additional retained samples after data collection and filtering. In the R2R-CE correction-allocation study, these budgets are approximately 190K state-level targets for $\mathcal{M}_{\mathrm{IA}}$, 11.5K selective action-supervised states for $\mathcal{M}_{\mathrm{AG}}$, or 200K action-supervised states for conventional DAgger.

\paragraph{Unified 7B baseline.}

The unified baseline uses Qwen2.5-VL-7B. Active-sub-instruction
and action targets form separate instruction-tuning samples.
Both use the global route instruction and eight RGB frames
uniformly sampled from the visual history, including the current
frame; neither the active sub-instruction nor execution state is
provided. At inference, the model directly predicts action chunks.

\section{Evaluation Protocols and Simulation Analyses}

\subsection{RxR-CE evaluation subset.}
The RxR-CE Val-Unseen split contains 3,669 episodes. All
99 affected episodes belong to scene
\texttt{oLBMNvg9in8}, where the start and goal positions lie
on different connected components of the provided navigation
mesh. Habitat therefore cannot compute a finite geodesic
path, resulting in infinite navigation error and undefined
SPL. We exclude these episodes for all methods and report
RxR-CE results on the same remaining 3,570 episodes.

\subsection{$\mathcal{M}_{\mathrm{IA}}$ Off-Path Evaluation Data and Protocol}

\paragraph{Controlled off-path history construction.}

For each aligned semantic segment, we initiate a perturbation from an interior expert state. We construct four model-independent perturbation types: (1) heading deviations of $30^\circ$, $45^\circ$, $60^\circ$, or $90^\circ$; (2) lateral detours that rotate by $\pm45^\circ$ and move forward by 0.25--0.75\,m; (3) short backtracking trajectories that first rotate by $180^\circ$ and then move 0.25--0.75\,m; and (4) fixed local zigzag loops consisting of 11 or 13 navigation primitives. A candidate perturbation is rejected if it causes a collision, crosses the current semantic waypoint, or prematurely reaches the current target.

After perturbation, an oracle controller guides the agent back to the endpoint pose of the same semantic segment, using a positional tolerance of 0.25\,m and a heading tolerance of $10^\circ$. The ground-truth segment index remains fixed throughout deviation and reconnection, so the active sub-instruction does not change during this process. Only observations collected during perturbation and reconnection are included in the evaluation; normal expert-path observations are excluded. This procedure produces 11,779 FG-R2R and 23,549 Landmark-RxR off-path queries. Representative examples are shown in Fig.~\ref{fig:mia_direct}.
\begin{figure*}[t]
    \centering
    \setlength{\tabcolsep}{3pt}

    \begin{tabular}{@{}ccc@{}}
        \includegraphics[width=0.3\textwidth]
        {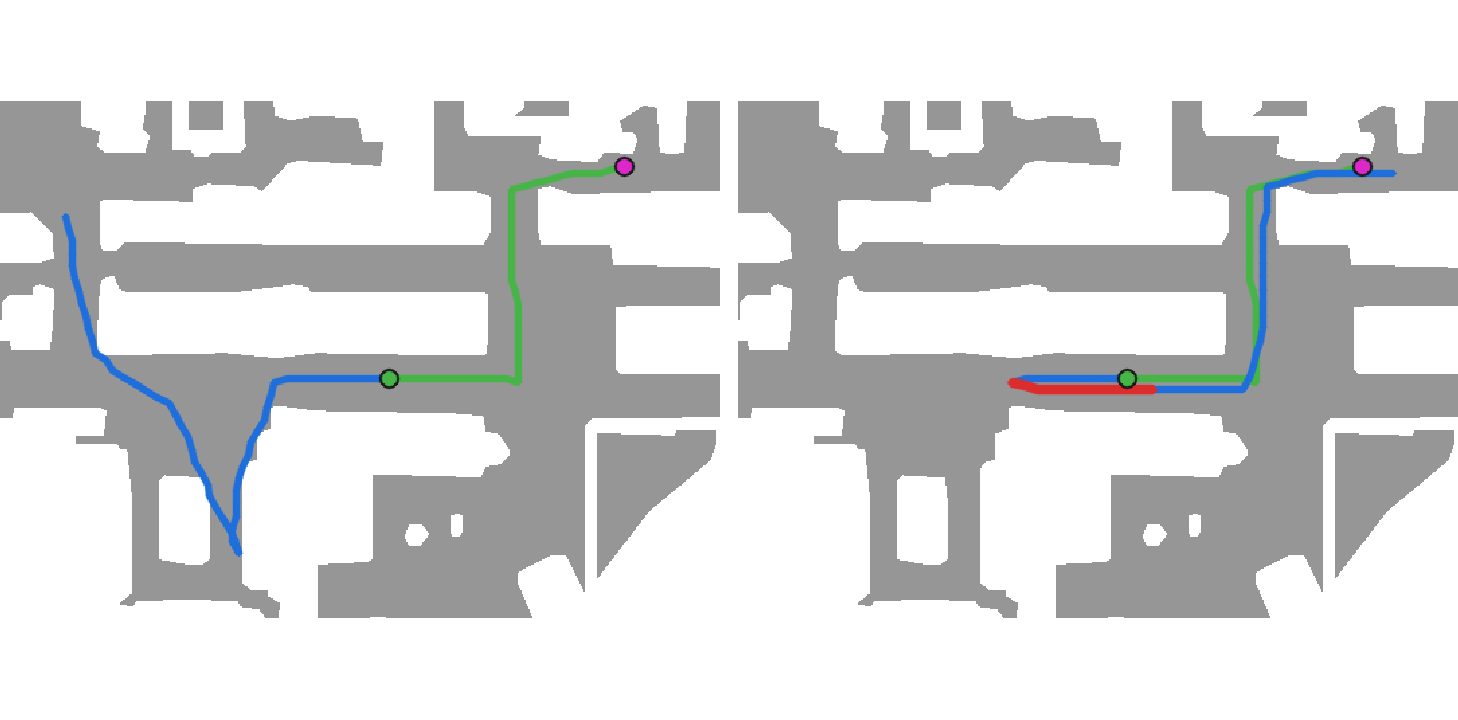}
        &
        \includegraphics[width=0.3\textwidth]
        {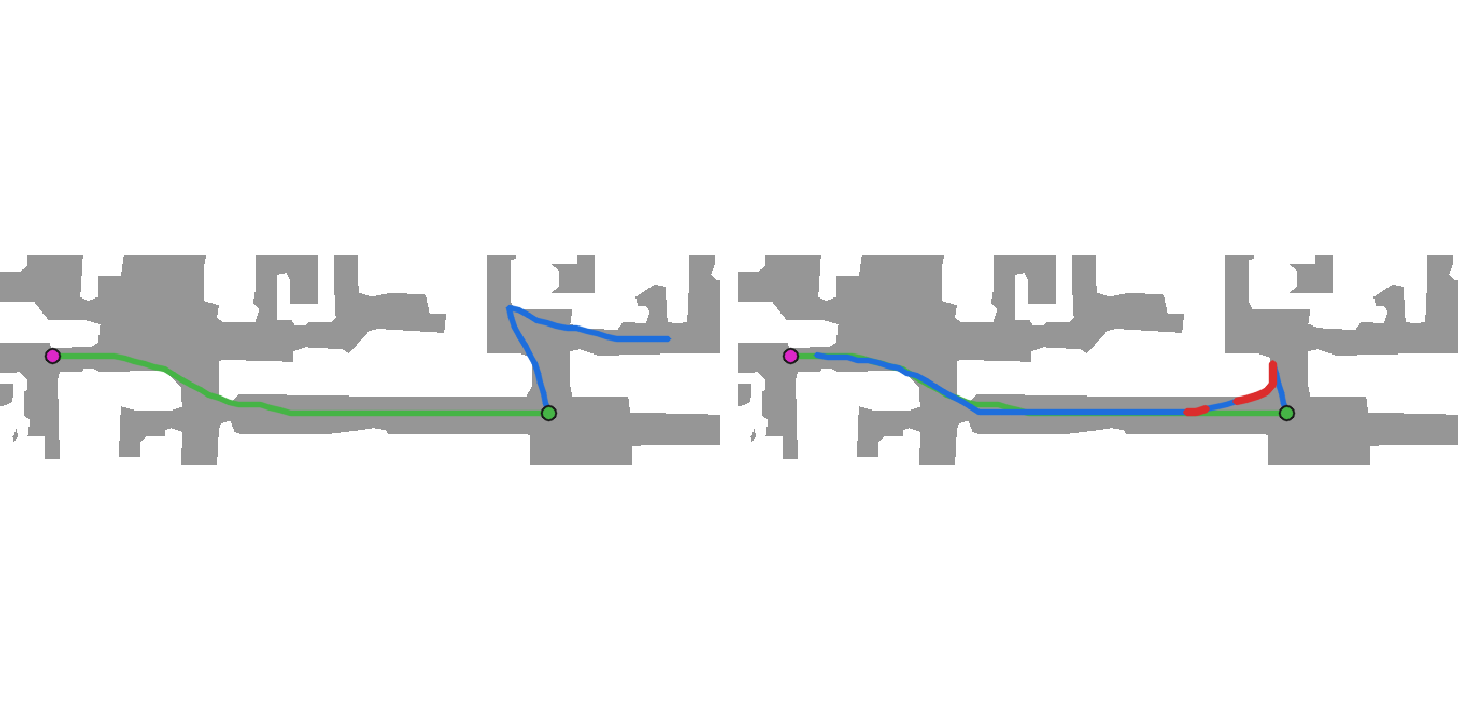}
        &
        \includegraphics[width=0.3\textwidth]
        {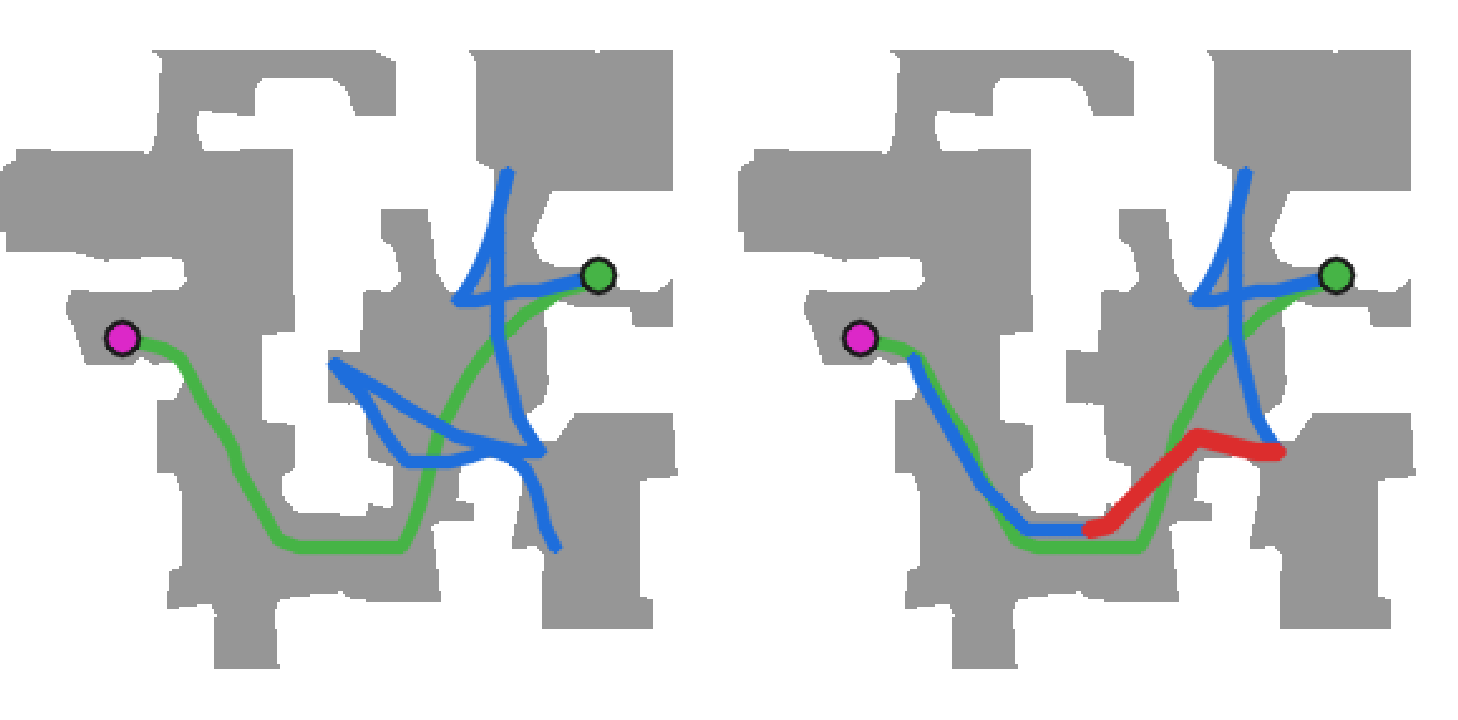}
        \\[-1pt]

        {\scriptsize (a)}
        &
        {\scriptsize (b)}
        &
        {\scriptsize (c)}
        \\[5pt]

        \includegraphics[width=0.3\textwidth]
        {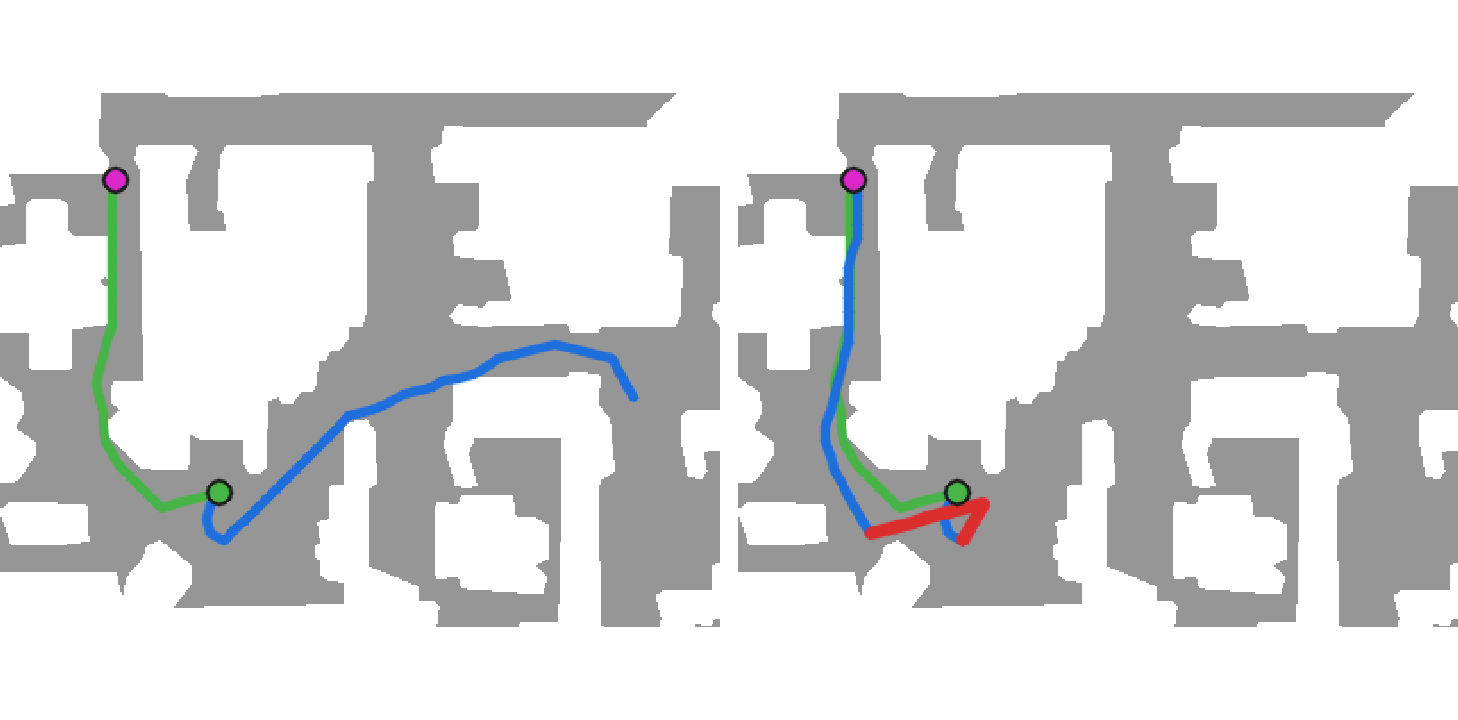}
        &
        \includegraphics[width=0.3\textwidth]
        {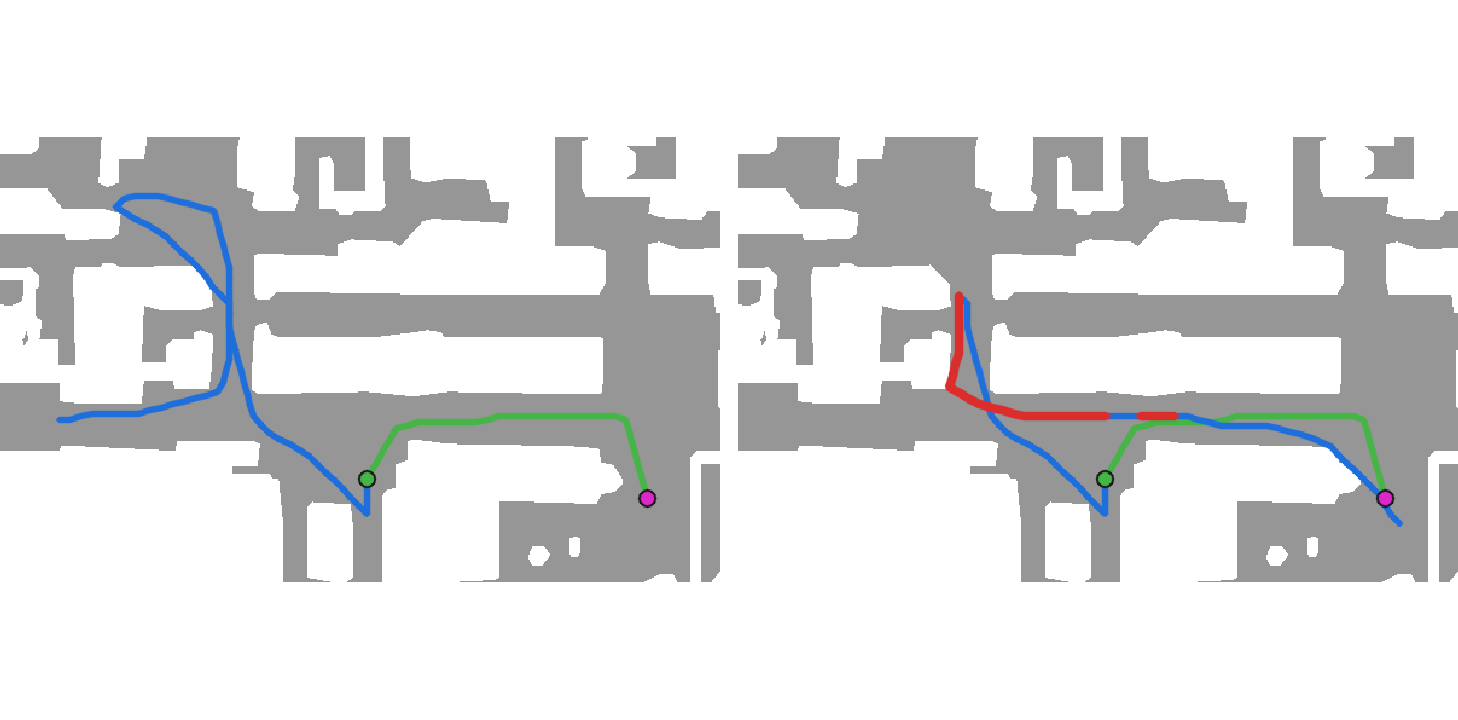}
        &
        \includegraphics[width=0.3\textwidth]
        {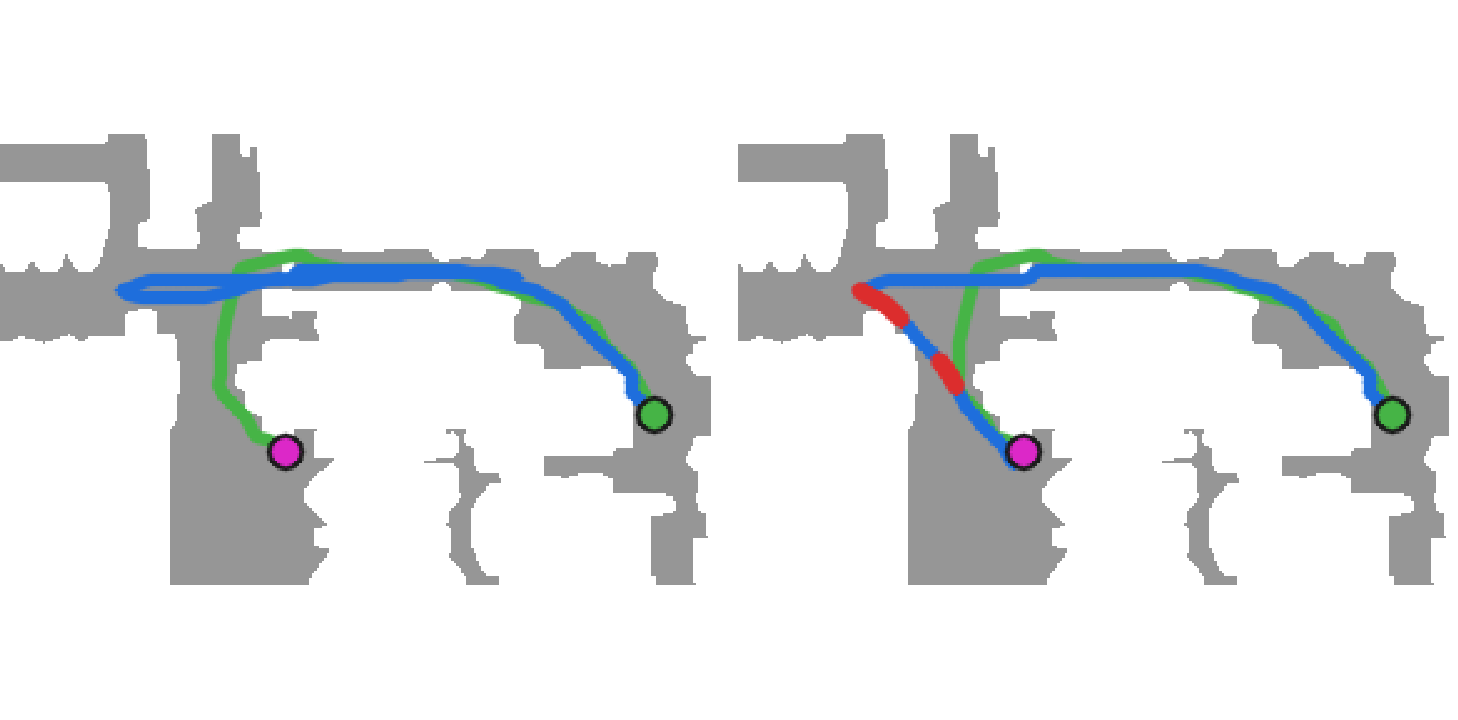}
        \\[-1pt]

        {\scriptsize (d)}
        &
        {\scriptsize (e)}
        &
        {\scriptsize (f)}
    \end{tabular}

\caption{
Representative recovery-routing cases. For each case, the
left panel shows execution with recovery disabled by forcing
the state to \texttt{Normal}, while the right panel shows the
same episode with predicted recovery routing enabled. Blue
and red segments denote actions executed under
\texttt{Normal} and \texttt{Recovering}, respectively, while
green denotes the reference trajectory.
}
    \label{fig:recover_topdown}
\end{figure*}

\paragraph{Evaluation protocol.}

We follow the strict active-sub-instruction evaluation
protocol described in the main paper. Because
$\mathcal{M}_{\mathrm{IA}}$ generates free-form text, each
prediction is matched to the most semantically similar
sub-instruction from the same route using cosine similarity
between text embeddings produced by
\texttt{all-MiniLM-L6-v2}. A query is counted as correct
when the top-$1$ matched sub-instruction is the aligned
ground-truth active step. Ground-truth step labels are used
only for evaluation and are not provided to the model during
inference.

\subsection{Recovery Routing and Cross-Policy Transfer}

\paragraph{Simulation analysis of recovery routing.}

\label{app:recover_analysis}

We conduct a paired episode-level analysis of predicted
execution-state routing on R2R-CE Val-Unseen. As illustrated
in Fig.~\ref{fig:recover_topdown}, we define the
\emph{triggered subset} using the Predicted Recovering
variant: an episode is included if
$\mathcal{M}_{\mathrm{IA}}$ predicts \texttt{Recovering}
at least once during navigation. This yields a fixed subset
of 791 out of 1,839 episodes. Fixed Normal and Predicted
Recovering are then evaluated on exactly the same episode
identifiers using the same checkpoint and evaluation
protocol.

\begin{table}[t]
\centering
\caption{
Paired recovery-routing analysis on R2R-CE Val-Unseen.
The triggered subset contains episodes where Predicted Recovering
outputs \texttt{Recovering} at least once. Transitions are measured
from Fixed Normal to Predicted Recovering.
}
\label{tab:app_recover_analysis}

\small
\setlength{\tabcolsep}{6pt}
\renewcommand{\arraystretch}{1.12}

\begin{tabular}{lcc}
\toprule
Metric & Count & Rate \\
\midrule
Triggered episodes
& $791/1{,}839$
& $43.01\%$ \\
\midrule
Fixed Normal success
& $283/791$
& $35.78\%$ \\
Predicted Recovering success
& $320/791$
& $40.46\%$ \\
Net improvement
& $+37$
& $+4.68$ pp \\
\midrule
Success $\rightarrow$ Success
& $199$
& -- \\
Failure $\rightarrow$ Failure
& $387$
& -- \\
Failure $\rightarrow$ Success
& $121$
& -- \\
Success $\rightarrow$ Failure
& $84$
& -- \\
\bottomrule
\end{tabular}
\end{table}

On the triggered subset, Predicted Recovering improves SR
from 35.78\% to 40.46\%, corresponding to a gain of
4.68 percentage points. It converts 121 Fixed-Normal
failures into successes while changing 84 successes into
failures, producing a net gain of 37 episodes. Since the
overall evaluation gains 39 successful episodes, 37 of these
additional successes arise from the triggered subset. This
indicates that the benefit of execution-state routing is
concentrated on episodes where recovery is actually invoked.
\begin{table*}[t]
\centering
\caption{Per-task results of the real-world evaluation.
R2S denotes Route2Step, SV denotes StreamVLN, and
SV+MIA denotes StreamVLN augmented with the predicted active
sub-instruction.}
\label{tab:app_realworld_tasks}
\footnotesize
\setlength{\tabcolsep}{2.5pt}
\renewcommand{\arraystretch}{1.08}
\begin{tabular}{p{0.10\textwidth}p{0.25\textwidth}cccccc}
\toprule
ID & Environment / route summary & Words & Steps & Trials & R2S & SV & SV+$\mathcal M_{\rm IA}$ \\
\midrule
I-1 & lab-1 & 120 & 7 & 5 & 3/5 & 0/5 & 0/5 \\
I-2 & lab-2 & 78 & 8 & 5 & 2/5 & -- & --\\
O-1 & park-1 & 14 & 2 & 5 & 2/5 & -- & -- \\ 
O-2 & park-2 & 24 & 4 & 3 & 3/3 & -- & -- \\ 
O-3 & park-3 & 16 & 3 & 3 & 1/3 & -- & -- \\ 
O-4 & outdoor parking lot & 33 & 3 & 3 & 3/3 & -- & -- \\
O-5 & living area-1& 45 & 9 & 3 & 1/3 & -- & -- \\
O-6 & living area-2 & 27 & 5 & 3 & 2/3 & -- & -- \\
O-7 & café area & 63 & 4 & 3 & 2/3 & -- & -- \\
\midrule
Total & 2 indoor + 7 outdoor tasks & -- & -- & 33 & 19/33 & -- & --\\
\bottomrule
\end{tabular}
\end{table*}

\paragraph{Cross-policy interface injection.}

All external navigation policies are kept frozen. For each policy, the Base and $+\mathcal{M}_{\mathrm{IA}}$ variants use the same checkpoint, evaluation episodes, and inference configuration. At each policy re-planning round, the $\mathcal{M}_{\mathrm{IA}}$ prediction is appended to the original textual input as

\begin{quote}
\small\ttfamily
Current Sub-instruction: \{$\mathcal{M}_{\mathrm{IA}}$ output\}.
\end{quote}
The field is incorporated following each method's original context-management implementation, and no other component of the inference pipeline is modified.

All results are obtained from our local evaluation using the released checkpoints and inference code. Because the released NaVid and Uni-NaVid checkpoints do not exactly reproduce the numbers reported in their original papers, improvements are measured relative to their corresponding locally reproduced Base results.

\section{Real-World Deployment Details}
\label{app:realworld}

\begin{figure*}[t]
    \centering

    \includegraphics[width=0.9\textwidth]{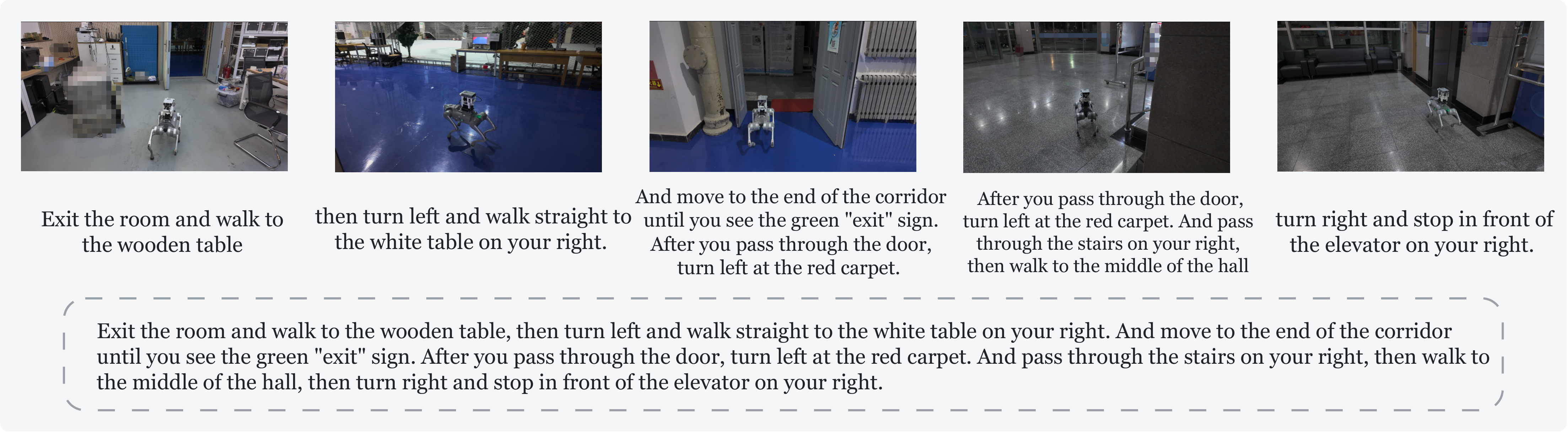}
    \vspace{-3pt}
    
    {\small\makebox[\textwidth][c]{(a) Lab-2}}

    \vspace{2pt}

    \includegraphics[width=0.9\textwidth]{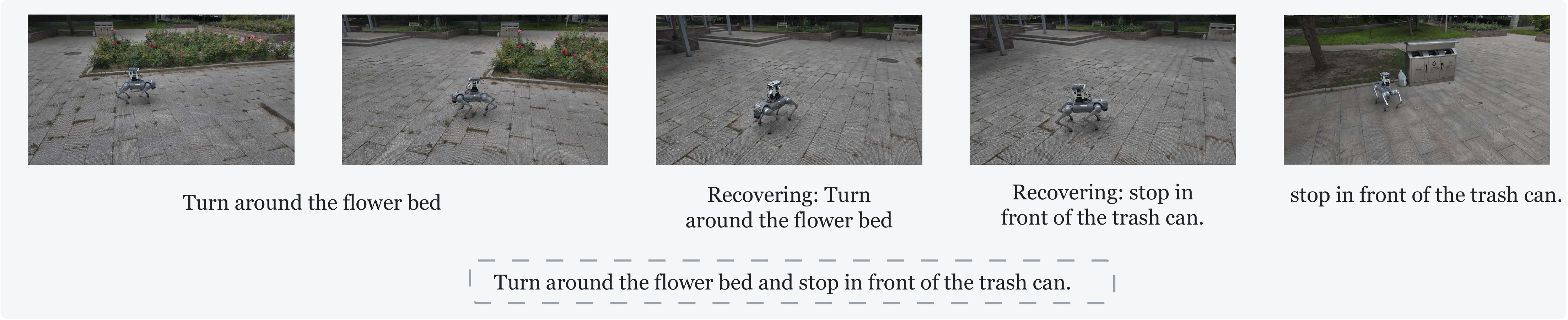}
    \vspace{-3pt}
    
    {\small\makebox[\textwidth][c]{(b) Park-1}}

    \vspace{2pt}

    \includegraphics[width=0.9\textwidth]{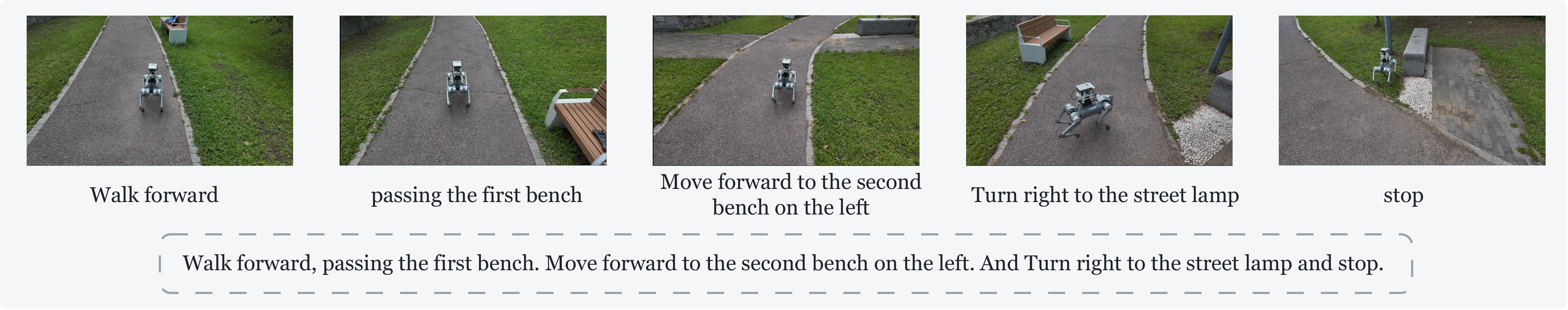}
    \vspace{-3pt}
    
    {\small\makebox[\textwidth][c]{(c) Park-2}}

    \vspace{2pt}

    \includegraphics[width=0.9\textwidth]{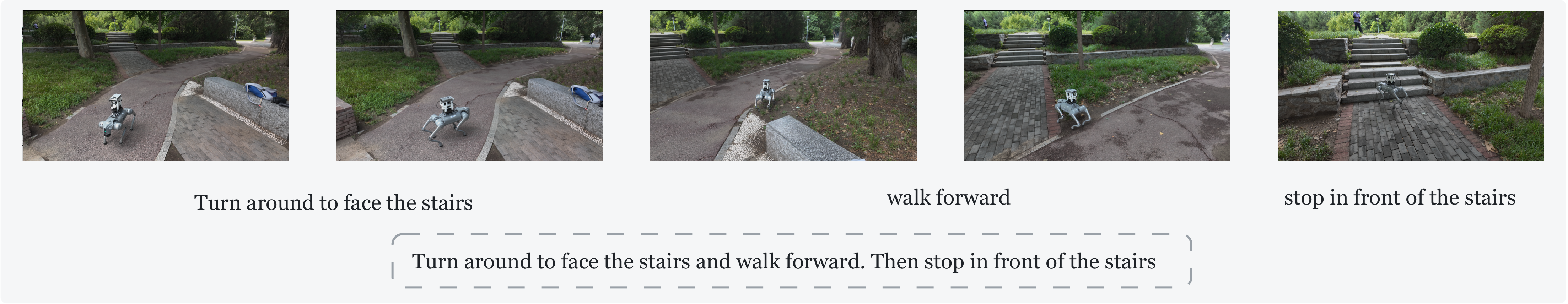}
    \vspace{-3pt}
    
    {\small\makebox[\textwidth][c]{(d) Park-3}}

    \vspace{2pt}

    \includegraphics[width=0.9\textwidth]{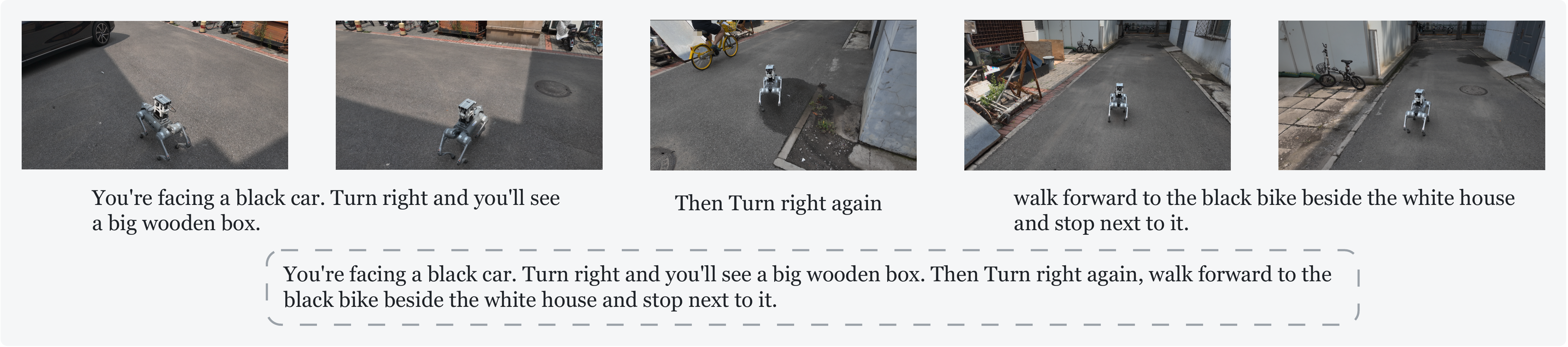}
    \vspace{-3pt}
    
    {\small\makebox[\textwidth][c]{(e) Outdoor Parking Lot}}

    \vspace{2pt}

    \includegraphics[width=0.9\textwidth]{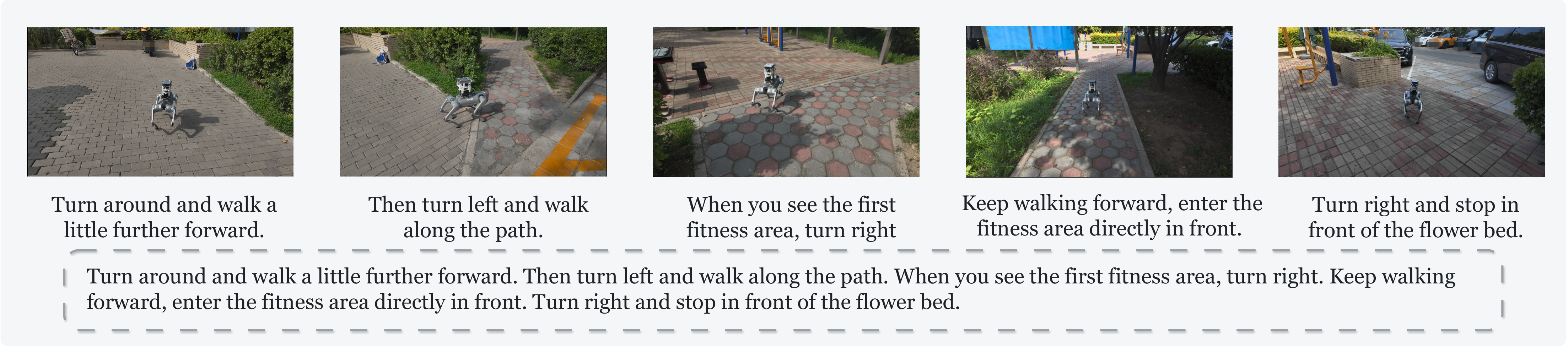}
    \vspace{-3pt}
    
    {\small\makebox[\textwidth][c]{(f) Living Area-1}}

    \caption{
    Real-world experiments (part I).
    Representative real-world navigation cases in diverse indoor and outdoor environments.
    }
    \label{fig:realworld_1}

\end{figure*}

\begin{figure*}[t]
    \centering

    \includegraphics[width=0.9\textwidth]{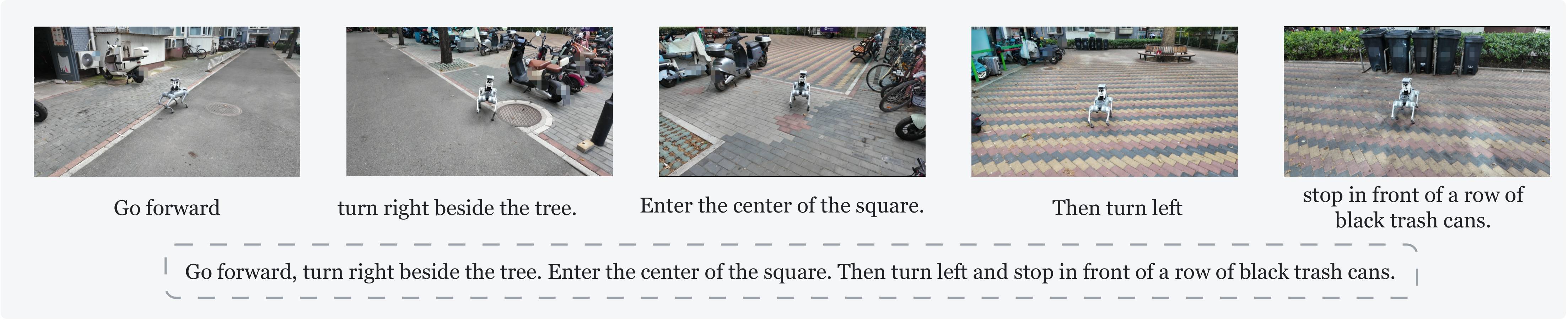}
    \vspace{1pt}
    
    {\small (g) Living Area-2}

    \vspace{4pt}

    \includegraphics[width=0.9\textwidth]{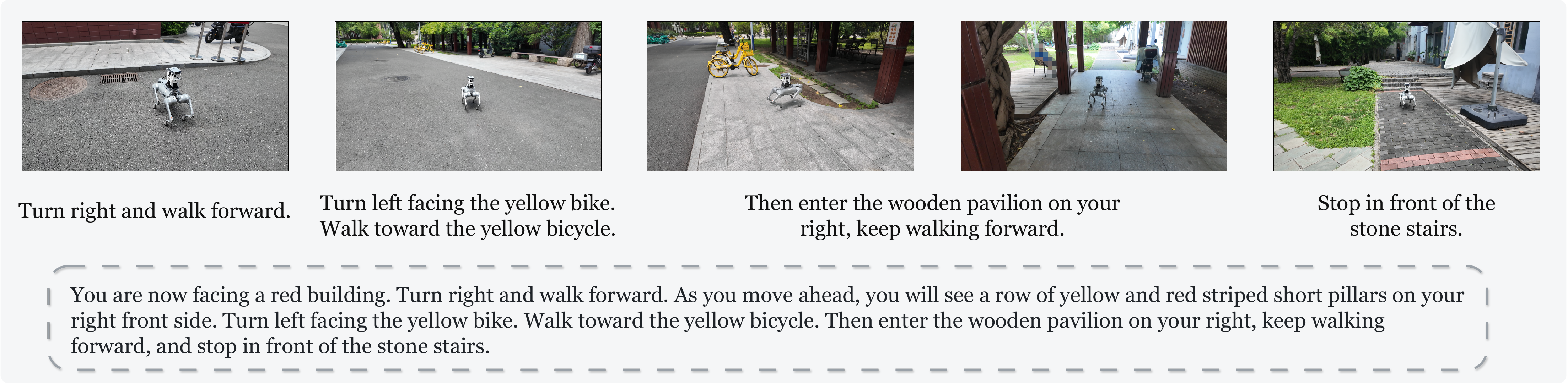}
    \vspace{1pt}
    
    {\small (h) Café area}

    \caption{
    Real-world experiments (part II).
    Additional representative real-world navigation cases.
    }
    \label{fig:realworld_2}
\end{figure*}

\begin{figure*}[b]
    \centering

    \includegraphics[width=0.9\textwidth]
    {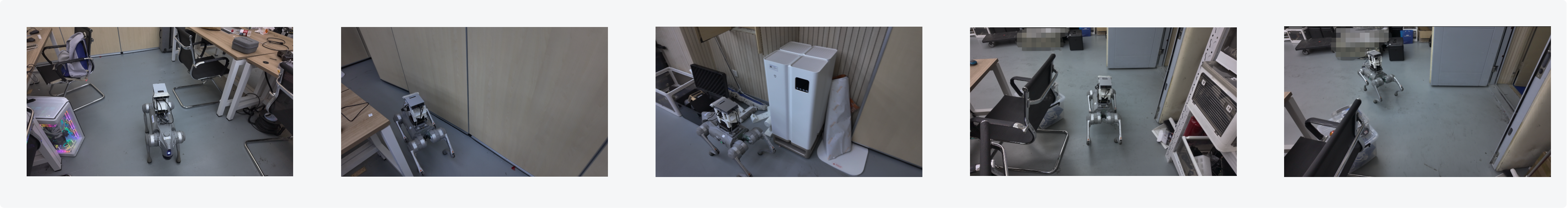}

    \vspace{1pt}
    
    {\small(a) StreamVLN}

    \vspace{5pt}

    \includegraphics[width=0.9\textwidth]
    {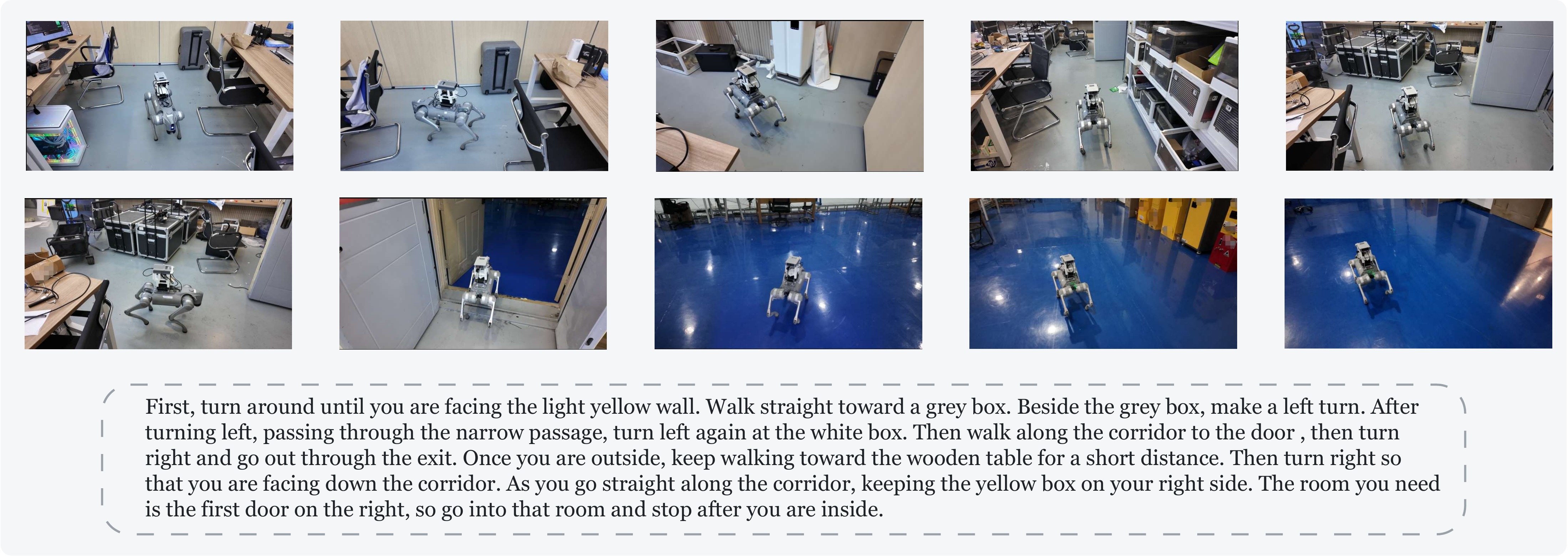}

    \vspace{1pt}
    
    {\small(b) StreamVLN + $\mathcal{M}_{\mathrm{IA}}$}
    \vspace{-5pt}
    \caption{
    Representative real-world comparison under the same complex
    indoor route instruction and initial pose.
    (a) StreamVLN deviates before completing the instructed route.
    (b) Augmenting the frozen StreamVLN policy with MIA provides
    the predicted active sub-instruction at each re-planning round,
    allowing the agent to maintain more instruction-consistent
    semantic progress, although the full route is not completed in
    the illustrated trial.
    }
    \label{fig:streamvln}
\end{figure*}

The real-world experiments use a Unitree GO2 quadruped
equipped with an Intel RealSense D455 camera. The navigation
models receive only monocular RGB observations and route
instructions and are transferred directly from simulation
without real-world fine-tuning. We evaluate nine indoor and
outdoor navigation tasks over 33 trials. A trial is successful
when the robot autonomously reaches and stops at the
predefined destination without external intervention.
Per-task statistics are reported in
Table~\ref{tab:app_realworld_tasks}. Representative navigation
cases are shown in Figs.~\ref{fig:realworld_1}
and~\ref{fig:realworld_2}, while Fig.~\ref{fig:streamvln} compares the frozen StreamVLN policy with its $+\mathcal{M}_{\mathrm{IA}}$ variant under the same complex indoor route instruction and initial pose.


\end{document}